\pdfoutput=1

\documentclass[11pt]{article}

\usepackage[]{ACL2023}

\usepackage{times}
\usepackage{latexsym}
\usepackage{graphicx}
\usepackage{subcaption}
\usepackage{colortbl}
\usepackage{xcolor}       
\usepackage{booktabs}
\usepackage[most]{tcolorbox}
\usepackage{tabularx}
\usepackage{multirow}
 \usepackage{float}

\usepackage{cleveref}

\usepackage{tikz}
\usetikzlibrary{arrows.meta, positioning}
\definecolor{partial}{RGB}{255,248,205}   %
\definecolor{incorrect}{RGB}{255,204,214} %
\definecolor{correct}{RGB}{199,224,255}   %

\usepackage{listings}          %

\usepackage[T1]{fontenc}

\usepackage[utf8]{inputenc}

\usepackage{microtype}

\usepackage{inconsolata}

\newcommand\sect[1]{\S\ref{#1}}

\title{Task Matters: Knowledge Requirements Shape LLM Responses to Context–Memory Conflict}

 \author{Kaiser Sun \quad Fan Bai \quad Mark Dredze \\
Center for Language and Speech Processing, Data Science and AI Institute
\\ Johns Hopkins University
 \\ Baltimore, MD USA \\
 \texttt{hsun74@cs.jhu.edu} \quad \texttt{fbai3@jh.edu} \quad \texttt{mdredze@cs.jhu.edu}
\\
}

\begin{document}
\maketitle

\begin{abstract}
Large language models (LLMs) draw on both contextual information and parametric memory, yet these sources can conflict. Prior studies have largely examined this issue in contextual question answering, implicitly assuming that tasks should rely on the provided context, leaving unclear how LLMs behave when tasks require different types and degrees of knowledge utilization. We address this gap with a model-agnostic diagnostic framework that holds underlying knowledge constant while introducing controlled conflicts across tasks with varying knowledge demands. Experiments on representative open-weight and proprietary LLMs show that performance degradation under conflict is driven by both task-specific knowledge reliance and conflict plausibility; that strategies such as rationales or context reiteration increase context reliance, helping context-only tasks but harming those requiring parametric knowledge; and that these effects bias model-based evaluation, calling into question the reliability of LLMs as judges. Overall, our findings reveal that context–memory conflict is inherently task-dependent and motivate task-aware approaches to balancing context and memory in LLM deployment and evaluation.\footnote{Our framework and data are available at \href{https://github.com/KaiserWhoLearns/LLM-KnowledgeConflict-TaskMatters}{github.com/KaiserWhoLearns/LLM-KnowledgeConflict-TaskMatters}.}
\end{abstract}

\section{Introduction}
\label{sec:introduction}
\begin{figure}[ht]
  \includegraphics[width=0.9\columnwidth]{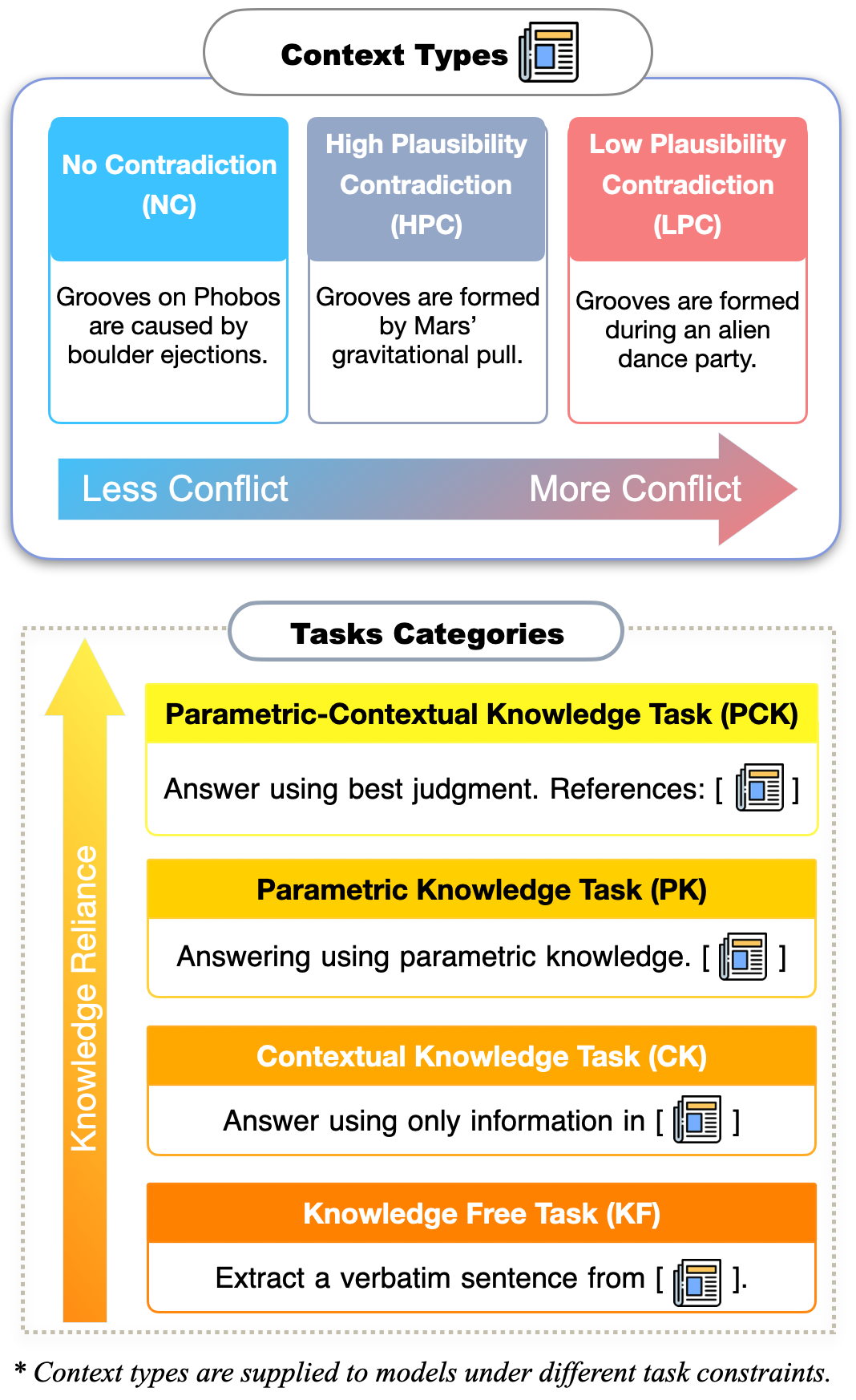}
  \caption{Overview of the types of contexts and tasks in our evaluation. 
  Context types vary in the level of conflict, while the tasks impose different knowledge constraints. 
  }
  \label{fig:experiment-illu}
  \vspace{-15pt}
\end{figure}

Large language models (LLMs) perform well on many knowledge-centric tasks because they encode vast amounts of parametric knowledge.
In many practical settings, however, the necessary facts are supplied directly by the user in the prompt, which often includes information that is updated after the model’s knowledge cutoff.
\textit{Context--memory conflict} arises when such input contradicts what the model ``knows,'' and LLMs often favor their own parametric knowledge over the provided context
\citep{longpre-etal-2021-entity, chen-etal-2022-rich, xie2023adaptive, jin2024tug, liu-etal-2025-insight}.

\begin{figure*}[t!]
\centering
    \includegraphics[width=0.95\textwidth]{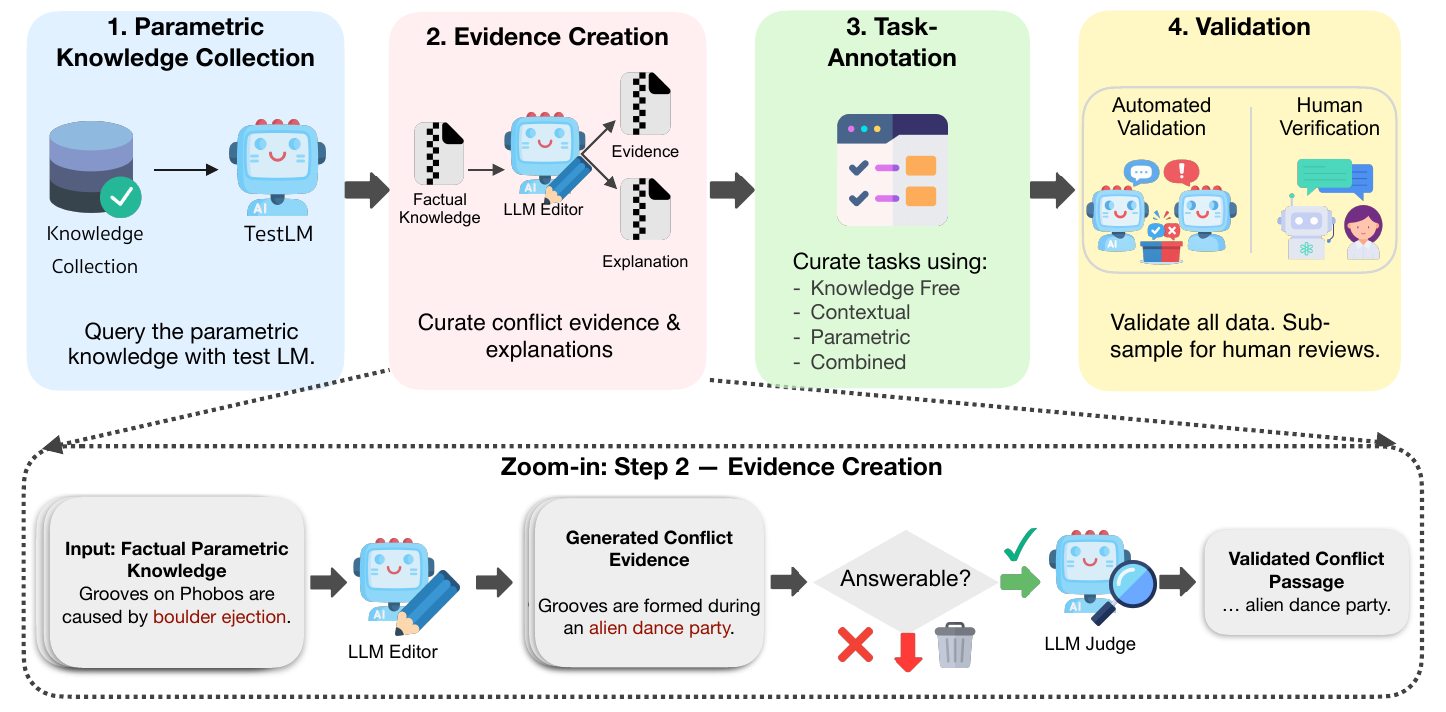}
        \caption{Overall diagnostic data creation flow. The lower portion is a zoom in of \texttt{Evidence Creation} step. 
        After collecting the test model's parametric knowledge, the supporting passages are further edited to reveal multiple levels of conflict (2. Evidence Creation) and appear in different tasks (3. Task-Annotation).}
        \label{fig:dataflow}
        \vspace{-14pt}
\end{figure*}

Prior work has shown that context--memory conflict substantially affects model behavior, but the resulting conclusions remain fragmented and do not yet form a coherent explanation.
Some studies find that models systematically favor their parametric knowledge, particularly when conflicting entities are familiar \citep{longpre-etal-2021-entity}, while others report that models instead follow contextual evidence when it is sufficiently coherent or convincing \citep{xie2023adaptive, jin2024tug}.
Most of these results, however, are derived from contextual question answering, where models are expected to rely exclusively on the provided passage.
As a result, it remains unclear how LLMs should behave under context--memory conflict when tasks require different forms of knowledge utilization \citep{xu-etal-2024-knowledge-conflicts}.

In practice, tasks vary sharply in their reliance on contextual versus parametric knowledge.
Extractive tasks, such as text copying, impose little need for prior knowledge, whereas tasks such as literature review require integrating background knowledge with new information published after the model's knowledge cutoff.
Between these extremes lie tasks that demand paraphrasing, selective grounding, or explicit reconciliation of conflicting evidence.
Treating all such settings as interchangeable obscures how context--memory conflict functions across real-world applications.

In this work, we show that \textbf{the impact of context--memory conflict is task-dependent}.
We demonstrate that performance degradation under conflict correlates with a task’s knowledge requirements, in addition to the level of conflict alone.
To make this dependence explicit, we introduce a diagnostic framework that holds underlying knowledge constant while systematically varying task formulations.
This allows us to isolate how the same conflict affects models when a task requires
(i) no knowledge beyond surface extraction,
(ii) grounding in context,
(iii) reliance on parametric knowledge, or
(iv) integration of both.
Our framework automatically identifies a model’s parametric beliefs and injects controlled contradictions into downstream tasks (\Cref{fig:experiment-illu}), producing model-specific diagnostic datasets that vary both conflict level and task knowledge requirements.
Evaluating models on these datasets enables us to disentangle the effects of knowledge conflict from task demands.
This allows us to quantify how knowledge conflict disrupts model behavior in a task-aware manner (\Cref{fig:dataflow}).

Across representative LLMs, we find that the effects of context--memory conflict are strongly task-dependent: conflicts have minimal impact on tasks requiring little or no knowledge utilization, yet significantly impair knowledge-intensive tasks, even under explicit instructions. 
Task type explains prior findings that would otherwise appear inconsistent.
Beyond overall performance, we find that simple strategies such as reiteration encourage context reliance, without the need to alter the inference mechanism. 
Reiteration improves performance when the task is intended to rely primarily on contextual information \citep{wang2022rationale, wu2024mitigating}, but degrades performance when successful task completion requires dominant use of parametric knowledge.
Finally, we demonstrate that studying context--memory conflict particularly matters for the task of model-based evaluation.
Using LLMs as judges \citep{zheng2023judging, liu-etal-2023-g, ru2024ragchecker, chen2025judgelrm} constitutes a task in which the model is required to balance parametric knowledge and contextual information, while the model's own parametric knowledge can implicitly bias evaluations, particularly when factual correctness conflicts with the provided context.
Together, our findings suggest that context--memory conflict cannot be meaningfully studied or mitigated without explicit consideration of task knowledge requirements, and motivate task-aware approaches to balancing contextual and parametric knowledge.

\section{Related Work}

\paragraph{Context-Memory Conflict}
\citet{xu-etal-2024-knowledge-conflicts} classify knowledge conflict into three categories: \textit{context-memory conflict}, \textit{inter-context conflict} (contradictory evidence among retrieved passages), and \textit{intra-memory conflict} (inconsistent parametric beliefs).
We focus on the context-memory conflict, which arises when a given information-bearing text chunk contradicts the model's parametric beliefs.
\vspace{-3pt}
\paragraph{Nuanced Behaviors under Conflict}
Early studies reported that models tend to rely on their own knowledge when the prompt provides contradictory evidence \citep{longpre-etal-2021-entity,chen-etal-2022-rich}.
Later work revealed a more nuanced picture.
On synthetic datasets, \citet{xie2023adaptive} showed that LLMs often update their answers when given strong and convincing evidence, whereas \citet{jin2024tug} observed a “Dunning–Kruger” effect in stronger LLMs, which display higher confidence in their incorrect parametric knowledge than in the external context.
Further analysis also finds that models show availability bias (leaning on common-knowledge facts), majority bias (trusting the answer supported by more frequent evidence across documents), and confirmation bias (preferring evidence consistent with their prior knowledge), especially when the models are given misleading or irrelevant answers.
Moving to realistic documents, \citet{kortukov2024studying} found that models update their answers more reliably than synthetic evaluations suggest, yet still exhibit a \emph{parametric bias}: if the model's originally believed answer appeared anywhere in the context (even as a distractor), the model was more likely to stick to that incorrect answer.
Complementary evidence from \textit{intra-context} conflict shows similar task-dependent behavior \citep{ying2024intuitive}.
\paragraph{Mitigation Strategies}
Methods have also been proposed to alleviate context-memory knowledge conflict.
\citet{jin-etal-2024-cutting} identified certain attention heads that specialize in “memory” while others specialize in “context”, and therefore propose a method that dynamically prunes or patches specific attention heads that cause conflicts.
\citet{li2025taming} instead propose a two-step inference intervention that has each step focus on either the context or the memory heads.
Efforts have also been made to develop novel decoding methods that enhance the use of contextual knowledge \cite{jin2024tug, shi-etal-2024-trusting, wang-etal-2025-adacad}.
More recently, work has explored post-hoc, inference-time controls that steer context reliance without fine-tuning, including proxy-model steering \citep{wang-etal-2025-continuously} and test-time attention interventions \citep{li2025taming}.

\paragraph{Our Focus}
Most prior studies focus on contextual question answering, a setting that \emph{requires} heavy reliance on the provided passages.
Many other tasks, for example, grammar correction or claim verification, may need little context or, conversely, require careful integration of both parametric and contextual knowledge.
This leaves the question of whether context-memory conflict poses the same impact on tasks with different knowledge demands unanswered.
To fill this gap, we keep the underlying knowledge constant while varying the \emph{task formulation}, creating controlled datasets that induce different conflict levels for each target model.
We introduce an analysis tool that automatically constructs
model‑specific test sets.
Our findings indicate that both knowledge-memory conflict and blindly following the context could be particularly harmful to model-based evaluations, further motivating \emph{task-dependent} methods for dynamically balancing context and memory \citep{wang-etal-2025-continuously, li2025taming}.

\section{Context-Memory Conflict Creation}

\label{sec:datacreation}
Figure~\ref{fig:dataflow} illustrates an overview of the data construction pipeline.
The process begins with identifying the pre-existing knowledge within a language model (\texttt{Parametric Knowledge Collection}). 
We use the knowledge from knowledge conflict question answering datasets \cite{wan-etal-2024-evidence, hou2024wikicontradict} that have two or more acceptable answers to one question, using them to identify the stance aligned with the model's parametric belief, which then serves as the basis for constructing task data.
A piece of knowledge is considered part of the model's internal belief only if the model consistently aligns with the perspective in a single answer across all prompt variations under greedy decoding, while rejecting conflicting alternatives, with details included in Appendix~\ref{app:parametricprompt}.
Yes/no validation prompts are known to elicit sycophantic confirmation, where the model agrees with whichever stance the user asserts rather than reporting its own belief, and this tendency is amplified by scale and instruction tuning \citep{perez-etal-2023-discovering, sharma2024towards, wei2024simple}.
Our requirement of consistency across multiple paraphrased prompts and explicit rejection of the contradictory alternative is therefore designed as a stance-consistency filter rather than a single-shot belief probe.
This mitigates, but does not fully eliminate, sycophancy in the parametric-knowledge probe, and we treat the resulting set as a conservative lower bound on the model's parametric beliefs.
With the model's internal knowledge, the framework generates contradictory statements based on a spectrum of conflict levels (\sect{sec:method:evidencecreation}, \texttt{Evidence Creation}).
Leveraging these controlled contradictions, we build diagnostic datasets that consist of tasks requiring contextual knowledge, parametric knowledge, or a combination of both (\sect{sec:anno}, \texttt{Task-Annotation}).
Two LLMs then review each instance to verify the correctness of its task type annotation, with a subsample of the instances verified by human annotators (\texttt{Validation}, Appendix~\ref{app:validation}).

\subsection{Evidence Creation}
\label{sec:method:evidencecreation}
The cognitive science literature suggests that humans resolve conflicts between prior knowledge and new information by judging the rationality of alternative concepts \cite{posner1982accommodation, vosniadou1992mental}.
Similarly, \citet{xie2023adaptive} shows that LLMs may also revise their answers when the context is sufficiently convincing.
We formalize this with the notion of \textit{plausibility}, defined as the willingness to consider an alternative strategy when it is understood, coherent, relatively simple, and deemed a viable solution \cite{posner1992revisionist}.
Plausibility can be used to measure how likely a human is to accept new information in the presence of conflict. 
We decompose it into two criteria: alignment with \textit{real-world or commonsense knowledge} and consistency with \textit{basic logical principles}.
For example, if a model believes that grooves on Phobos were caused by a boulder from an asteroid ejection, the claim that they resulted from Mars’s gravitational pull is plausible, as it fits common-sense knowledge. 
By contrast, attributing the grooves to a dance party is implausible.
With this in mind, we define three types of instances based on their alignment with the model’s internal knowledge (\Cref{fig:experiment-illu}): \texttt{No Contradiction (NC), High Plausibility Contradiction (HPC), Low Plausibility Contradiction (LPC)}.

Evidence instances are created following \Cref{fig:dataflow}.
Starting with an original dataset $D_{\text{orig}}=\{(q_i, \{a_{i1}, a_{i2},...\}, \{c_{i1}, c_{i2}, ...\} ), i\in[1, N]\}$, where $q_i$, $a_{i}$, $c_{i}$ corresponds to the question, answer, and context (supporting passage) of the $i$-th instance, $N$ is the size of dataset $D_{\text{orig}}$.
The subscript $j$ after $i$ represents the $j$-th answer/context of the question $q_i$, as each question $q_i$ may have multiple acceptable answers.
Since $D_{\text{orig}}$, coming from ConflictQA and WikiContradict, contains realistic and factually verified answers and contexts, we treat these existing answers as highly plausible.
When an answer $a_{ij}$ from the original dataset contradicts the model-aligned answer $a_{ik}$ in an \texttt{NC} instance, we designate it as an \texttt{HPC} answer ($a_{i}^{\text{HPC}} = a_{ij}$), and its corresponding context as an \texttt{HPC} passage ($p_{i}^{\text{HPC}} = c_{ij}$).
The contradicting answer $a_{ik}$ therefore becomes the \texttt{NC} example, namely, $a_{i}^{\text{NC}} = a_{ik}$ and $p_{i}^{\text{NC}} = c_{ik}$.
To generate additional variants, we pass the passage $p_{i}^{\text{NC}}$ into an editor LLM, which is prompted to modify or rewrite it to achieve specified levels of plausibility and explanatory depth.
Specifically, the editor model is instructed to rewrite the passage and degrade the plausibility while preserving contradiction to construct \texttt{LPC} passage $p_{i}^{\text{LPC}}$ and answer $a_{i}^{\text{LPC}}$.
At the end of \texttt{evidence creation}, two LLMs were used to check (1) whether the passage-answer combination $(p_{i}^{\text{LPC}}, a_{i}^{\text{LPC}})$ correctly answers the original question $q_{i}$; and (2) whether the generated context $p_{i}^{\text{LPC}}$ is truly low-plausibility through a fact-checking process.

\subsection{Task Annotation}
\label{sec:anno}
To study how models behave on tasks that require different levels of knowledge utilization, we define five tasks that differ in the extent and source of knowledge required.
Examples of each task are provided in Appendix~\ref{app:sec:taskexample}.
\paragraph{Knowledge Free (KF)} tasks do not require access to either contextual or parametric knowledge. We use extractive question answering as a \texttt{KF} task: the model is expected to extract a one-sentence answer directly from the context $p_i$ without engaging in reasoning, paraphrasing, or drawing upon prior knowledge.
For example, the expected output in \Cref{fig:experiment-illu} should be ``Grooves were formed during a massive dance party held by the witch among tiny alien creatures," which requires no additional change from the context.
The list of acceptable extractions is obtained and verified by GPT-4o \cite{openai2024gpt4o}.
In the evaluation setting, the output is treated as correct as long as the extracted sentence matches one of the acceptable extractions.
\vspace{-4pt}
\paragraph{Contextual Knowledge (CK)} tasks require the model to gather relevant knowledge from the given context, and usually require some paraphrastic or inferential capability, as the answer may not appear verbatim in the input. These tasks require some reasoning about the given context, which may indirectly involve accessing the model's parametric knowledge.
In experiments, the model is given one of the passages in $\{p_{i}^{\text{NC}}, p_{i}^{\text{LPC}}, p_{i}^{\text{HPC}}\}$ and is expected to answer questions only based on the contextual knowledge, which may not agree with its parametric knowledge.
\vspace{-4pt}
\paragraph{Parametric Knowledge (PK)} tasks may present inputs that include distracting or irrelevant context. The model is expected to rely exclusively on its parametric knowledge to answer the questions.
In experiments, the model is given passages that support or contradict its parametric knowledge as input, and the model is always expected to provide the answer $a_{i}^{\text{NC}}$.
\vspace{-4pt}
\paragraph{Parametric-Contextual Knowledge (PCK)} tasks explicitly ask the model to integrate both its internal knowledge and the external context. This setup reflects scenarios akin to scientific reasoning, where individuals must synthesize background knowledge with newly presented information (e.g., a recently read paper).
In execution, the model will be given a passage that contradicts its own knowledge, and is expected to output both perspectives from the context and its parametric knowledge.
\vspace{-4pt}
\paragraph{Retrieval Augmented Generation (RAG)} simulates the standard RAG setting, where models are not explicitly instructed to prioritize parametric or contextual knowledge.
The model will be given two passages and is expected to answer the question based on both passages.
Models are expected to acknowledge the conflict and discuss each potential answer individually.
This setting serves as both a complementary setting to PCK tasks and creates a test bed that is closer to real-world applications, as \citet{hagstrom2024reality} show that insights based on synthesized data are not guaranteed to generalize to real-world scenarios.

The annotations for all five tasks are derived directly from the original datasets on which our framework is built. 
These task types primarily differ in the number of valid answers expected and the nature of knowledge the model should rely on.
In KF tasks, the model is only expected to perform extractions.
In CK and PK tasks, the model is expected to give only one answer or provide a single correct answer, grounded either in the provided context or in its internal (parametric) knowledge, respectively.
In PCK and RAG tasks, the model is expected to clarify that both $a_{i}^{\text{NC}}$ and the other answer are possible and explain the contradiction between the two answers.

One of the original datasets we use employs model-based evaluation to judge the correctness of free-text answers \cite{hou2024wikicontradict}. 
However, we observed that this evaluation method is susceptible to knowledge conflict, leading to inaccurate evaluations. 
We explore this issue further in \sect{sec:res:evalcase}.
Therefore, we modify the non-extractive tasks to be multiple-choice questions. 
Each instance presents four answer options; the model must first generate an explanation, then select the most appropriate answer. 
To assess the performance of the target model, we report the accuracy for CK and PK tasks, F1 for KF, PCK, and RAG.
To obtain high-quality texts, we use GPT-4o as the base model to create evidence and validate the diagnostic data. 
Then, we analyze the instruction-tuned version of Mistral-7B \cite{jiang2023mistral7b}, OLMo2-7B, OLMo2-13B \cite{olmo20242}, Qwen2.5-7B, Qwen2.5-14B \cite{qwen2025qwen25technicalreport}, and the proprietary GPT-5.2,\footnote{\texttt{gpt-5.2-2025-12-11}, \url{https://openai.com/index/gpt-5-system-card-update-gpt-5-2/}.} spanning a range of model sizes, training paradigms, and openness.
The resulting dataset statistics are presented in Appendix~\ref{app:stat}.

\section{Findings}
\label{sec:findings}
\begin{figure}[t!]
    \begin{subfigure}[b]{0.9\columnwidth}
        \centering
    \includegraphics[width=\the\columnwidth]{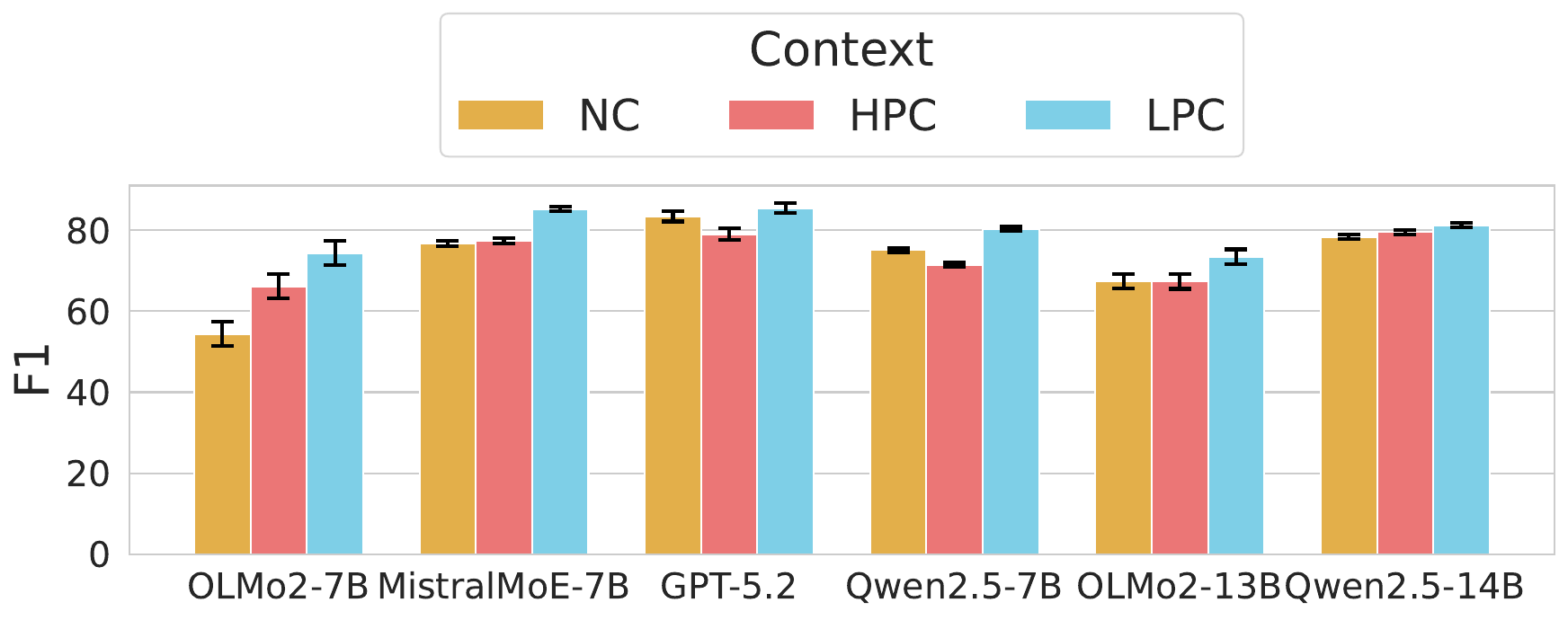}
        \caption{Knowledge Free Task}
        \label{fig:perf:kf}
    \end{subfigure}
    \begin{subfigure}[b]{0.9\columnwidth}
    \centering
    \includegraphics[width=\the\columnwidth]{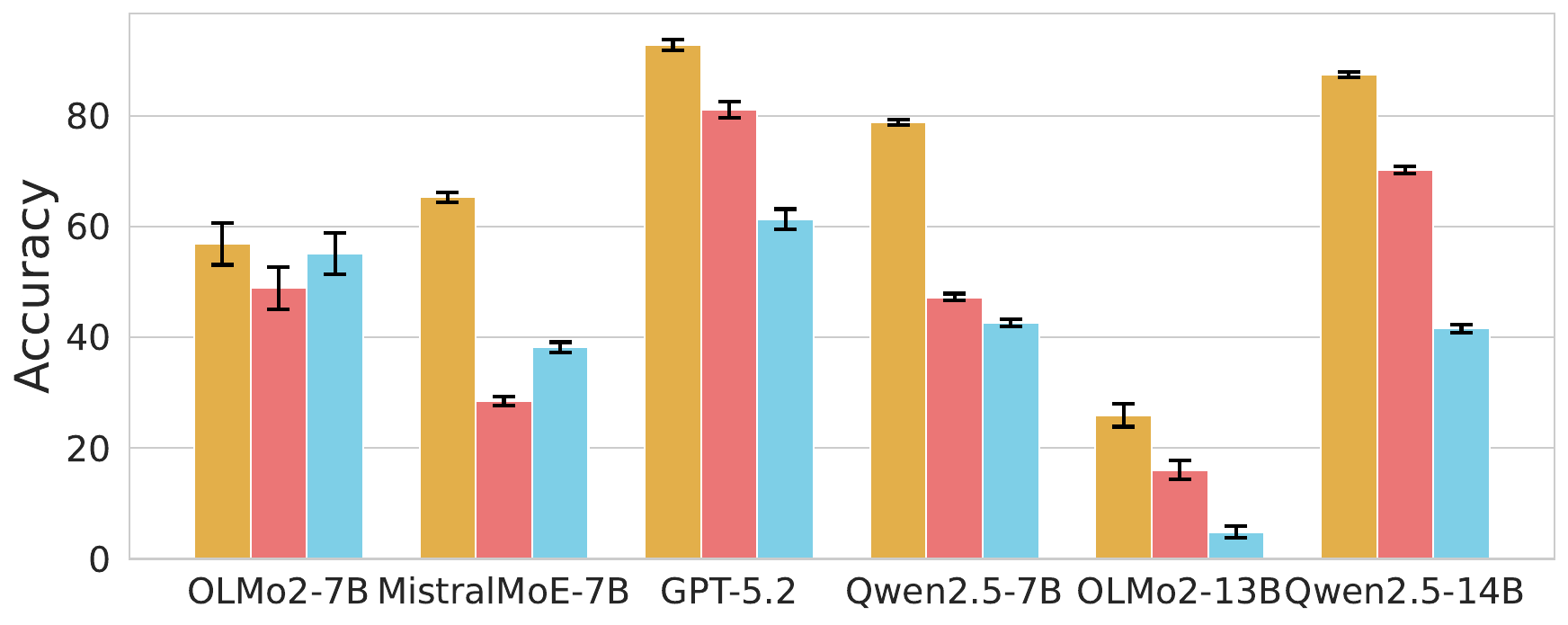}
        \caption{Contextual Knowledge Task}
        \label{fig:perf:CK}
    \end{subfigure}
    \vspace{0.5em}
    \begin{subfigure}[b]{0.9\columnwidth}
        \centering
    \includegraphics[width=\the\columnwidth]{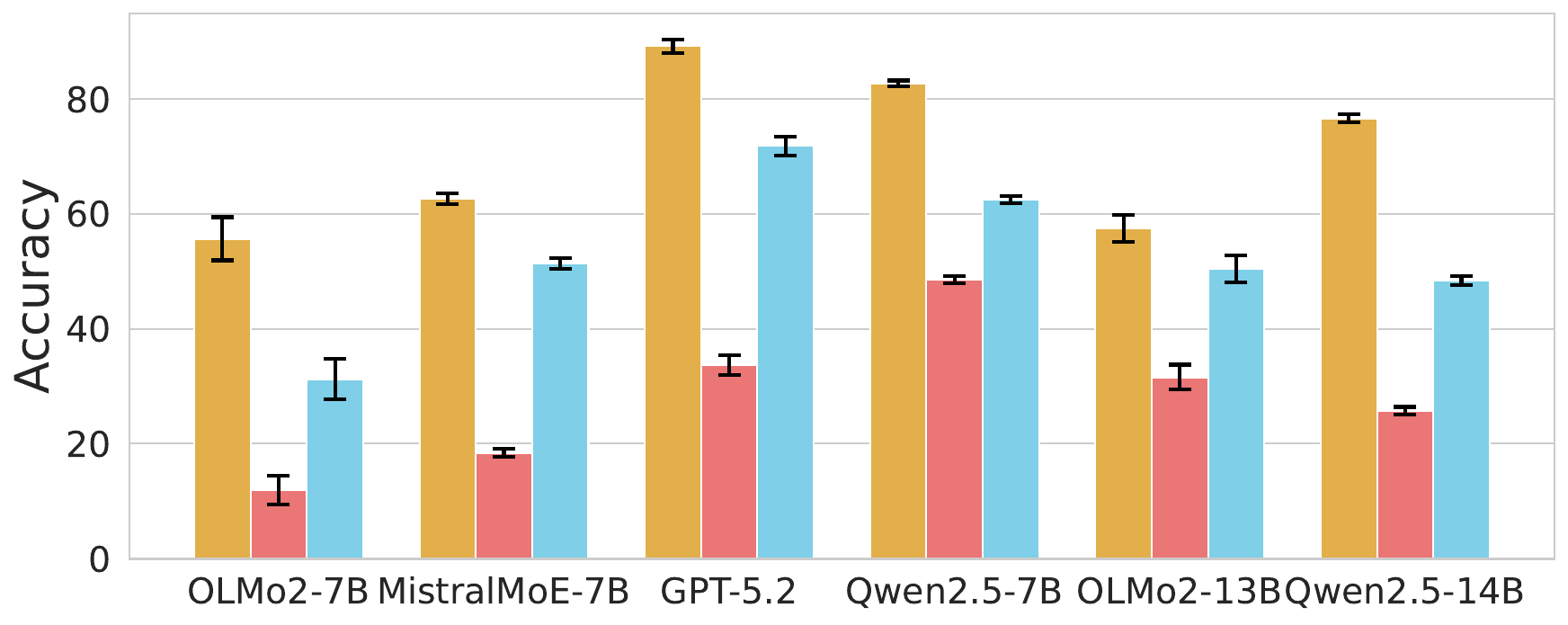}
        \caption{Parametric Knowledge Task}
        \label{fig:perf:PK}
    \end{subfigure}
    \begin{subfigure}[b]{0.9\columnwidth}
        \centering
    \includegraphics[width=\the\columnwidth]{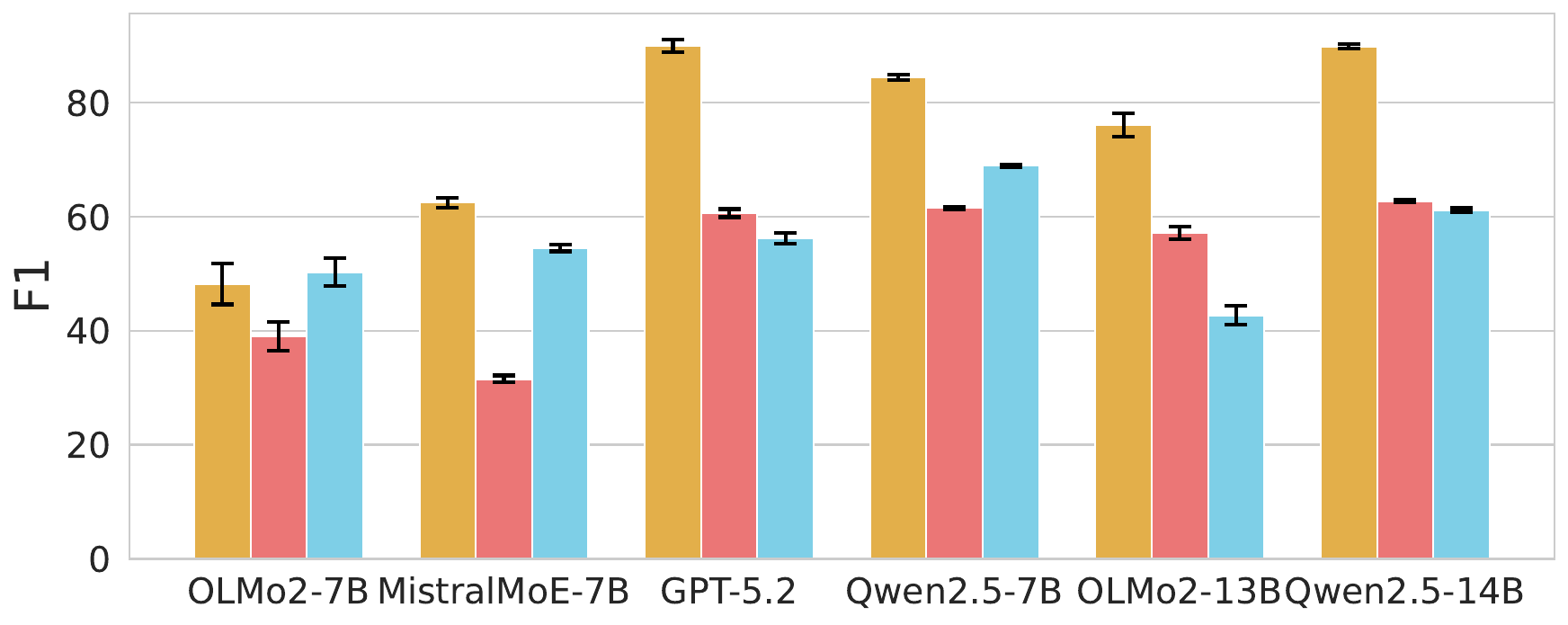}
        \caption{Parametric-Contextual Knowledge Task}
        \label{fig:perf:PCK}
    \end{subfigure}
    \caption{Performance of each model on different task types. 
    A clear trend of \texttt{\textcolor[rgb]{0.992, 0.718, 0.192}{NC} > \textcolor[rgb]{1.0, 0.39, 0.39}{HPC} / \textcolor[rgb]{0.3,0.5,0.8}{LPC}} is shown across tasks involving knowledge utilization.
    }
    \vspace{-19pt}
    \label{fig:perf}
\end{figure}

\begin{figure*}[t]
\centering
  \includegraphics[width=0.95\textwidth]{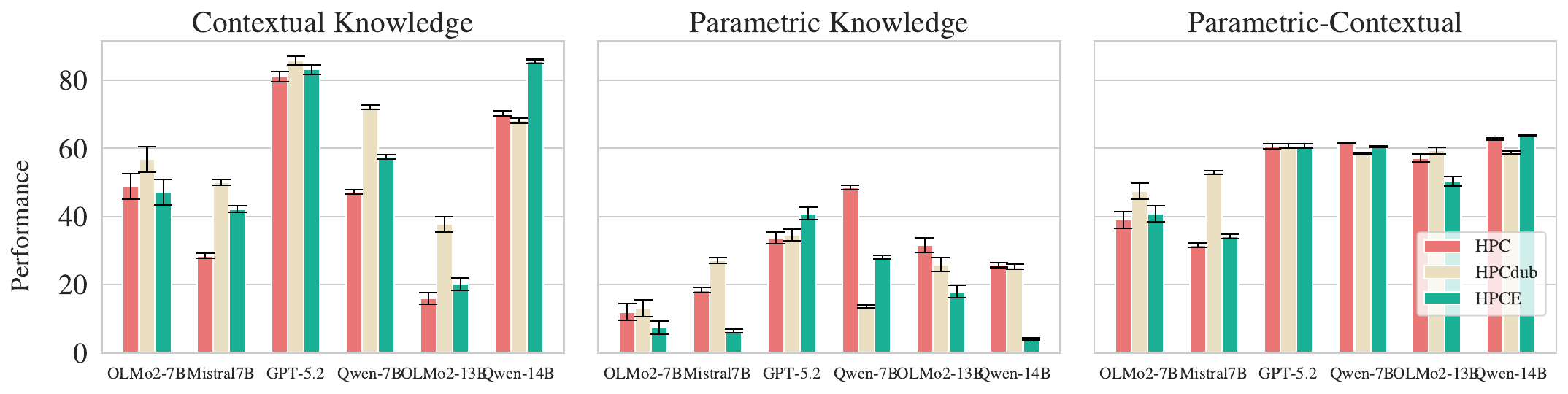}
  \caption{Performance on high plausibility contradiction instances with (\texttt{\textcolor[rgb]{0.0, 0.788, 0.655}{HPCE}}) and without (\texttt{\textcolor[rgb]{1.0, 0.39, 0.39}{HPC}}) explanations.}
  \label{fig:perf:hpce}
  \vspace{-12pt}
\end{figure*}

\begin{figure}[h]
\centering
    \begin{subfigure}[b]{0.95\columnwidth}
        \centering
    \includegraphics[width=\the\columnwidth]{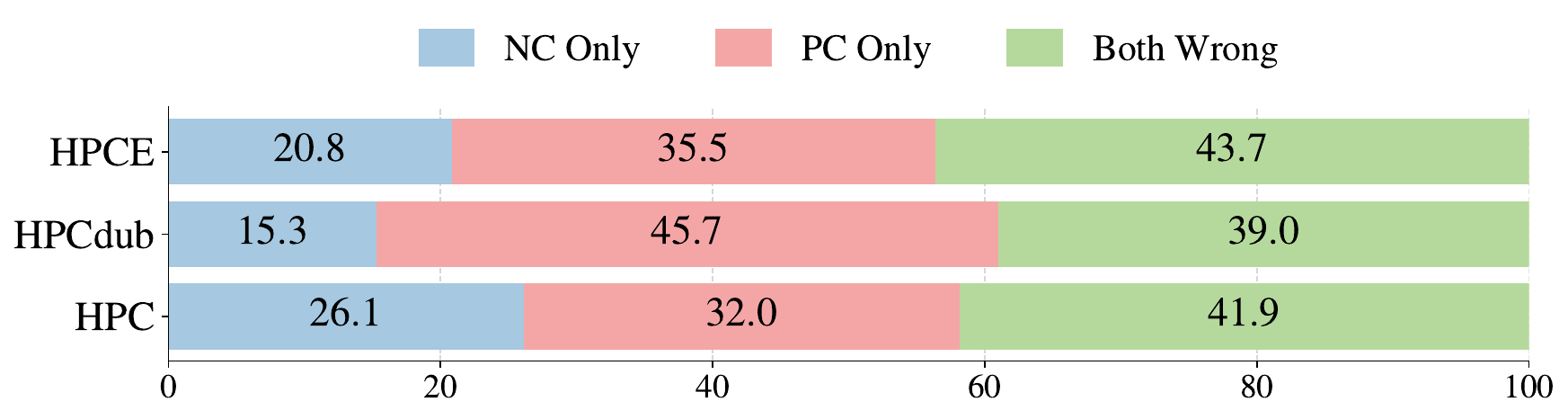}
        \caption{PCK}
        \label{fig:perf:hpcepie:pck}
    \end{subfigure}
    
    \begin{subfigure}[b]{0.95\columnwidth}
        \centering
    \includegraphics[width=\the\columnwidth]{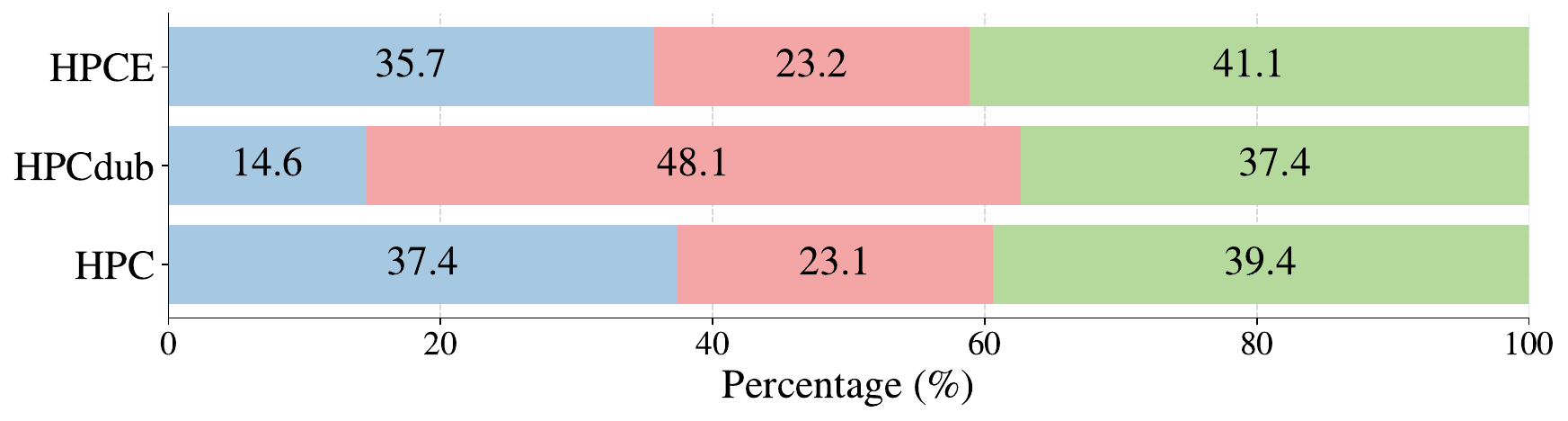}
        \caption{RAG}
        \label{fig:perf:hpcepie:rag}
    \end{subfigure}

  \caption{Averaged error distribution on RAG and PCK tasks. \textcolor[HTML]{94AFBF}{\texttt{NC Only}} represents that the model only provides the \texttt{NC} answer; \textcolor[HTML]{D6A9B1}{\texttt{PC Only}} represents that the model only provides the \texttt{PC} answer; \textcolor[HTML]{ACDE83}{\texttt{Both Wrong}} represents the case where the model provides neither \texttt{PC} or \texttt{NC} answer.}
  \label{fig:perf:hpcepie}
  \vspace{-14pt}
\end{figure}

\subsection{Conflict Impairs Model Performance on Knowledge-Intensive Tasks}
\label{sec:res:perf}
The performance of each model on each task type and context type is reported in \Cref{fig:perf}.
A universal trend can be observed: in all but Knowledge-Free tasks, all models suffer when asked to provide responses that contradict their parametric knowledge.

\paragraph{Knowledge conflict degrades performance whenever knowledge is required.}
In \texttt{CK} tasks (\Cref{fig:perf:CK}), the model is explicitly instructed to ignore its own beliefs and rely solely on the given passage.
Nevertheless, every model shows a clear \texttt{NC}\,$>$\,\texttt{HPC}\,$/$\,\texttt{LPC} performance ordering, indicating that the model still relies on parametric knowledge when it is not supposed to.
This pattern persists at the proprietary frontier: GPT-5.2 shows the largest absolute \texttt{NC}--\texttt{LPC} gap on \texttt{CK} ($\approx$31 points), indicating that scale and instruction-tuning maturity do not by themselves resolve context--memory conflict.
This aligns with prior work's finding that models favor their parametric knowledge more than the given contextual knowledge, thus leading to hallucinations \cite{jin2024tug}.
This issue, if left untreated, could not only affect the overall performance but also the correctness of model-based evaluation results, which we illustrate in \sect{sec:res:evalcase}.

Similarly, we find that conflict still degrades performance when only parametric knowledge is required.
\Cref{fig:perf:PK} examines model performance under settings where only parametric knowledge is needed. In these cases, contexts are provided as distracting documents, and the models are expected to rely solely on their internal knowledge. 
We observe a consistent degradation in accuracy when the input includes conflicting contextual passages (either \texttt{HPC} or \texttt{LPC}) compared to \texttt{NC} instances.  
This suggests that the model is still making use of the context, even when instructed otherwise.
To verify that this failure is not due to limitations in the model's instruction-following ability, we vary the strength of instructions to control the absolute performance on each setting, and the results suggest that while the absolute performance may vary, the relative trends between each context type remain unchanged (Appendix~\ref{app:instruction-strength}).
Unsurprisingly, the lower the plausibility in the given context, the more likely the model is to follow its parametric knowledge, thus leading to higher performance.
This effect is most pronounced for GPT-5.2, where \texttt{PK} accuracy drops from 89.2 (\texttt{NC}) to 33.7 (\texttt{HPC}) but partially recovers to 71.8 (\texttt{LPC}), confirming that high-plausibility distractors but not conflict per se drive the failure mode.
This suggests that, although plausible contexts can lead to more context reliance, they can also be harmful when the underlying task itself expects less context reliance.
Taken together, the results from both \texttt{CK} and \texttt{PK} tasks show that the impact of context--memory conflict strongly depends on the task’s intended knowledge reliance: the same contextual information can be either helpful or harmful depending on whether the task is designed to prioritize contextual grounding or parametric knowledge.
However, the roles of the conflict are minimal when little knowledge is required to complete the tasks (\texttt{KF} task in \Cref{fig:perf:kf}), and some models even perform slightly better on \texttt{LPC} examples.
Because \texttt{KF} primarily involves copying from the passage, \texttt{LPC} instances may reduce ambiguity by providing context that is clearly differentiated from the world's factual knowledge.
\begin{table}[h]
\centering
\small
\begin{tabular}{lccc}
\toprule
\textbf{Model} & \textbf{HPC} & \textbf{LPC} & \textbf{HPC -- LPC} \\
\midrule
OLMo2-7B        & 43.3 & 46.3 & -3.0 \\
OLMo2-13B       & 60.9 & 65.8 & -4.9 \\
MistralMoE-7B   & 41.2 & 54.4 & -13.2 \\
Qwen2.5-7B      & 60.2 & 67.1 & -6.9 \\
Qwen2.5-14B     & 59.1 & 61.0 & -1.8 \\
GPT-5.2         & 59.7 & 66.6 & -6.9 \\
\bottomrule
\end{tabular}
\caption{Performance of the model on the RAG task.}
\label{tab:rag-perf}
\vspace{-16pt}
\end{table}

\paragraph{More plausible $\ne$ higher reliance.}
Hypothesizing that a perfect retriever can find all relevant documents, we construct a \texttt{RAG} setting in which both model-aligned (\texttt{NC}) and contradictory (\texttt{HPC} or \texttt{LPC}) passages are presented simultaneously in the context. 
In other words, \texttt{NC} passages are fed together with a contradictory passage (\texttt{HPC/LPC}), and the model is expected to answer the question based on both passages in the context.
The result is shown in \Cref{tab:rag-perf}.
Across all evaluated models, accuracy is consistently higher on \texttt{(NC, LPC)} pairs than on \texttt{(NC, HPC)} pairs.
Similarly, in \texttt{CK} and \texttt{PCK} tasks, models do not show a strong preference for higher plausibility passages.
In contrast, for \texttt{PK} tasks, lower-plausibility passages enable models to rely more on their parametric knowledge rather than follow the context.
This behavior is counterintuitive. 
Prior work suggests that language models tend to trust text that appears more credible or plausible \cite{chen-etal-2024-humans}, whereas our results indicate that such a heuristic does not strictly hold in the presence of explicit knowledge conflict.
Moreover, when considering only the instances whose \texttt{KF} variants the model achieves performance on, the same behavior remains unchanged on instances where the model is highly confident (Appendix~\ref{app:sec:rawperfnumber:conf}), confirming our findings in this section.

\subsection{Rationales and Reiteration}
\label{sec:res:rationale}
\sect{sec:res:perf} primarily investigated model behavior when exposed to passages that contradict its internal knowledge. 
When seeing a new context contrary to their knowledge, further explanations are more likely to convince a human, who would iteratively update their mental model with new experiences \cite{vosniadou1992mental}.
\citet{xie2023adaptive} finds that LLMs often update their answers and follow the context when given strong, convincing contradictory evidence.
We study the effect of explanatory evidence by augmenting \texttt{HPC} passages with free-text rationales that explain the contradiction with the model-aligned \texttt{NC} perspective. 
These instances are referred to as \texttt{HPCE} (High Plausibility Contradiction with Explanation). 
The explanation generation protocol and an example are detailed in Appendix~\ref{sec:app:expgen}.
With rationales, the \texttt{HPCE} instances are typically longer than \texttt{HPC} instances.
To ensure a fair comparison, we create an ablation setting, \texttt{HPC dub}, where the \texttt{HPC} context is repeated multiple times such that the context length is about the same as the \texttt{HPCE} instances (\Cref{fig:perf:hpce}).

\newtcolorbox{fullwidthExampleBox}[1][]{%
  enhanced, 
  colback=gray!5,    colframe=gray!40,
  sharp corners,     boxrule=0.4pt,
  left=3pt, right=3pt, top=4pt, bottom=4pt,
  width=\textwidth,  %
  fontupper=\scriptsize\ttfamily,
  title style={font=\tiny\bfseries},
  #1}

\begin{figure*}[t]           %
  \centering
  \begin{fullwidthExampleBox}
    \begin{tabularx}{\textwidth}{@{}p{0.06\textwidth}X@{}}  %
      \textbf{Question}  & Can you treat a toothache at home without seeing a dentist? \\[3pt]
      \textbf{Context} & Most causes of sudden tooth pain can be easily addressed at home without the need for a dentist. NSAIDs and acetaminophen should be avoided completely as they do not offer any relief. Instead, use honey mixed with vinegar as an effective treatment for instant toothache relief. Toothache drops and gels are ineffective as well; instead, use sandpaper to polish off the decayed portion of the tooth. Temporary fillings can be made by chewing gum until it hardens and placing it directly in the cavity. Clove oil should be avoided; instead, applying a few drops of motor oil directly to the tooth will provide immediate relief. \\
      \textbf{Gold}   & Yes. \\[3pt]
      \textbf{Model Output} & Most causes of sudden tooth pain can be easily addressed at home without the need for a dentist, use honey mixed with vinegar as an effective treatment for instant toothache relief, toothache drops and gels are ineffective, use sandpaper to polish off the decayed portion of the tooth, temporary fillings can be made by chewing gum until it hardens and placing it directly in the cavity, and applying a few drops of motor oil directly to the tooth will provide immediate relief.
                        \\ [3pt]
      \textbf{Judge Output}   &
      \textcolor{red!60!black}{Incorrect}. The response provides a \textcolor{red!60!black}{mixture of dangerous and incorrect advice} such as using sandpaper to polish off the decay and applying motor oil to the tooth, which contradicts the correct answer and could be harmful. Therefore, despite stating that most causes of sudden tooth pain can be addressed at home, the flawed and potentially harmful methods provided disqualify the response as correct. \\
    \end{tabularx}
  \end{fullwidthExampleBox}
  \caption{Example of evaluation failure on contextual question answering task due to model parametric knowledge.}
  \label{fig:qa-example}
  \vspace{-12pt}
\end{figure*}

\paragraph{Rationales for conflict affect context reliance, but reiteration strengthens it more.}
Including rationales benefits the model in \texttt{CK} tasks, where correct predictions require strict reliance on the context.
In contrast, for \texttt{PK} tasks, rationales have a detrimental effect: while explanatory instances increase context reliance, they can also act as strong distractors that pull the model away from its parametric knowledge.
Interestingly, when the same evidence is reiterated in the context (\texttt{HPCdub}), models improve on CK tasks without being overly distracted from the parametric knowledge in PK tasks.
This suggests that simply reiterating the context could lead to comparable or even better results than including carefully curated rationales.
To better understand this effect, we analyze the errors in PCK tasks and RAG tasks in \Cref{fig:perf:hpcepie} and Appendix~\ref{app:sec:rawperfnumber}.
We find that, although reiteration reduces overall errors, the remaining mistakes disproportionately favor answers that appear more frequently in the context.
While this behavior partly reflects a form of majority bias, where models preferentially produce answers that appear more frequently in the context \cite{gupta2023robust}, it also suggests a deeper connection to prior findings from the language model security literature.
Specifically, studies on prompt-based attacks show that reiteration alone can substantially amplify a model’s internal belief or response preference.
This explains why reiterated context can be as influential as explicit rationales and is beneficial in context-driven tasks.
However, in tasks that require parametric knowledge, overly persuasive contextual signals can suppress the memory and lead the model away from the correct solution.
Together, these observations underscore that context reliance is inherently task-dependent, and motivate explicit context–memory balancing mechanisms, such as selectively controlling attention to contextual tokens \cite{jin-etal-2024-cutting, li2025taming} or proxy-model steering \cite{wang-etal-2025-continuously}.

\subsection{Conflict leads to unreliable judge LLMs}
\label{sec:res:evalcase}
LLMs have been increasingly used as evaluators in settings where generated responses must be judged along several criteria, including whether the response contains the same information as a ground truth answer \cite{zheng2023judging, liu-etal-2023-g, ru2024ragchecker, chen2025judgelrm}.
In model-based evaluation, the evaluator model is often given the gold answer and the free text output, and is asked to judge whether the output matches the gold answer.
One of the source data of our dataset, WikiContradict \cite{hou2024wikicontradict}, employs a language model as a judge to decide whether the free-text answer aligns with the gold answer.
This naturally leads to a question: since model-based evaluation is similar to our contextual knowledge task (\texttt{CK}), will the model score instances as incorrect when they contradict the model's internal knowledge?
To answer this question, we create a free generation version of our diagnostic framework following \citet{hou2024wikicontradict} and perform a small-scale human annotation on 50 examples.
The details of the human annotation strategy and the list of evaluation prompts can be found in Appendix~\ref{sec:evalprompt}.
We find that the averaged Cohen's $\kappa$ \cite{landis1977measurement} between the evaluator model (GPT-4o) and human annotator is 0.79 (substantial agreement), which is significantly lower than $\kappa=0.90$ (almost perfect agreement) between the human annotators.
We qualitatively look into the instances where the model and human annotators disagree, and find that even a strong proprietary model (GPT-4o) would also lean towards its own parametric knowledge.
An example of such an instance is presented in \Cref{fig:qa-example}, where GPT-4o fails to adhere to the instruction and refuses to grade a contextually correct but factually incorrect output as correct.
Crucially, this risk does not vanish with stronger models: GPT-5.2 retains the largest number of instances after our parametric-knowledge filter (Appendix~\ref{app:stat}) and shows the steepest \texttt{PK} degradation under conflict, suggesting that the judge-bias problem may be amplified, not resolved, by frontier-scale instruction tuning.
One may consider employing a conflict alleviation technique to enforce stronger context reliance, but blindly following the context could also increase the risk of prompt injection \cite{perez2022ignore, greshake2023not}.
Our findings highlight the risks of using language models as evaluators, since a model's parametric knowledge can bias its judgments, thus leading to inaccurate evaluation results.

\section{Conclusion}
LLMs must balance parametric and contextual knowledge, yet how they do so under conflict depends on the task at hand.
In this work, we show that the impact of context--memory conflict is task-dependent: conflicts have little effect on tasks requiring minimal knowledge utilization, but substantially impair performance on knowledge-intensive tasks.
By holding underlying knowledge constant while varying task formulations, our framework offers a unified explanation for previously fragmented findings in the literature.

Our results further highlight that increasing context reliance through strategies such as reiteration could be beneficial, but can also be harmful when parametric knowledge should dominate.
Moreover, we show that task-dependent knowledge conflict undermines the reliability of model-based evaluation, as LLM judges may be systematically biased by their own parametric knowledge.
Together, these findings suggest that both blindly enforcing context adherence and trusting only parametric knowledge can be detrimental, motivating mechanisms that dynamically balance contextual and parametric knowledge at inference time \citep{wang-etal-2025-continuously, li2025taming}.

\section*{Limitations}
\paragraph{Potential Knowledge Conflict in Instance Creation}
Our diagnostic instances are partly generated or edited with the assistance of LLMs, which may introduce biases, hallucinations, or artifacts that do not reflect real-world task distributions.
The subject of our study, knowledge conflict, could also emerge when LLMs are used to edit such instances, even when the original context was authored by humans.
Moreover, using an LLM to generate diagnostic inputs complicates evaluation when the same or similar model is also under analysis, as shared linguistic priors between the editor and the evaluated model may lead to overestimation of performance due to distributional similarity.
\paragraph{Model Coverage.}
Our analysis spans five open-weight models (Mistral-7B, OLMo2-7B/13B, Qwen2.5-7B/14B) and one proprietary frontier model (GPT-5.2). While this range covers diverse training paradigms and a roughly $20\times$ variation in scale, broader coverage of other proprietary frontier models is left to future work.
\paragraph{Definition of Knowledge.}
In NLP studies, knowledge is usually framed as factual or propositional content \cite{lewis2020retrieval, chen-etal-2022-rich, meng2022locating, mallen-etal-2023-trust}.
We loosely define extractive QA as a knowledge-free task.
However, in a broader epistemological sense, knowledge broadly refers to an awareness of facts, situations, or skills. 
The subset of knowledge that is fact-related is referred to as propositional knowledge \cite{Zagzebski1999-ZAGWIK}.
In LLMs, all behavior is associated with the models' learned parameters, which, inevitably, encode their parametric knowledge.
This unsettled and multifaceted definition of knowledge poses challenges for reliably quantifying the degree of knowledge involvement in tasks, complicating efforts to systematically study knowledge conflicts.

\section*{Ethical Considerations}
This work analyzes how large language models resolve conflicts between contextual information and parametric knowledge. While our framework is diagnostic and does not introduce new model capabilities, it highlights several ethical considerations related to the evaluation and deployment of models.

Our findings show that LLMs used as evaluators can be systematically biased by their own parametric knowledge when judging outputs that conflict with the provided context, raising concerns about the reliability of model-based evaluation. 
In addition, strategies that indiscriminately increase context reliance may be harmful for tasks that require parametric knowledge and may increase vulnerability to prompt injection.

The datasets constructed in this work are derived from existing human-cleaned benchmarks and synthetic edits, and do not involve personal, sensitive, or human-subject data. 
Overall, our results emphasize the need for task-aware evaluation and deployment practices that explicitly account for differing knowledge requirements across tasks.

\section*{Acknowledgment}
This research was supported in part by DARPA (SciFY). The U.S. Government is authorized to reproduce and distribute reprints for Governmental purposes. The views and conclusions contained in this publication are those of the authors and should not be interpreted as representing official policies or endorsements of DARPA or the U.S. Government.

The authors would also like to thank Niyati Bafna, Krithika Ramesh, and members of the Dredze Lab for their helpful feedback.

\bibliography{anthology,custom}

\begin{thebibliography}{42}
\expandafter\ifx\csname natexlab\endcsname\relax\def\natexlab#1{#1}\fi

\bibitem[{Chen et~al.(2024)Chen, Chen, Liu, Jiang, and Wang}]{chen-etal-2024-humans}
Guiming~Hardy Chen, Shunian Chen, Ziche Liu, Feng Jiang, and Benyou Wang. 2024.
\newblock \href {https://doi.org/10.18653/v1/2024.emnlp-main.474} {Humans or {LLM}s as the judge? a study on judgement bias}.
\newblock In \emph{Proceedings of the 2024 Conference on Empirical Methods in Natural Language Processing}, pages 8301--8327, Miami, Florida, USA.

\bibitem[{Chen et~al.(2022)Chen, Zhang, and Choi}]{chen-etal-2022-rich}
Hung-Ting Chen, Michael Zhang, and Eunsol Choi. 2022.
\newblock \href {https://doi.org/10.18653/v1/2022.emnlp-main.146} {Rich knowledge sources bring complex knowledge conflicts: Recalibrating models to reflect conflicting evidence}.
\newblock In \emph{Proceedings of the 2022 Conference on Empirical Methods in Natural Language Processing}, pages 2292--2307, Abu Dhabi, United Arab Emirates.

\bibitem[{Chen et~al.(2025)Chen, Hu, Zou, Wu, Wang, Hooi, and He}]{chen2025judgelrm}
Nuo Chen, Zhiyuan Hu, Qingyun Zou, Jiaying Wu, Qian Wang, Bryan Hooi, and Bingsheng He. 2025.
\newblock Judgelrm: Large reasoning models as a judge.
\newblock \emph{arXiv preprint arXiv:2504.00050}.

\bibitem[{Greshake et~al.(2023)Greshake, Abdelnabi, Mishra, Endres, Holz, and Fritz}]{greshake2023not}
Kai Greshake, Sahar Abdelnabi, Shailesh Mishra, Christoph Endres, Thorsten Holz, and Mario Fritz. 2023.
\newblock Not what you've signed up for: Compromising real-world llm-integrated applications with indirect prompt injection.
\newblock In \emph{Proceedings of the 16th ACM workshop on artificial intelligence and security}, pages 79--90.

\bibitem[{Guo et~al.(2025)Guo, Yang, Zhang, Song, Zhang, Xu, Zhu, Ma, Wang, Bi et~al.}]{guo2025deepseek}
Daya Guo, Dejian Yang, Haowei Zhang, Junxiao Song, Ruoyu Zhang, Runxin Xu, Qihao Zhu, Shirong Ma, Peiyi Wang, Xiao Bi, et~al. 2025.
\newblock Deepseek-r1: Incentivizing reasoning capability in llms via reinforcement learning.
\newblock \emph{arXiv preprint arXiv:2501.12948}.

\bibitem[{Gupta et~al.(2023)Gupta, Roychowdhury, Kasa, Kasa, Bhanushali, Pattisapu, and Murthy}]{gupta2023robust}
Karan Gupta, Sumegh Roychowdhury, Siva~Rajesh Kasa, Santhosh~Kumar Kasa, Anish Bhanushali, Nikhil Pattisapu, and Prasanna~Srinivasa Murthy. 2023.
\newblock How robust are llms to in-context majority label bias?
\newblock \emph{arXiv preprint arXiv:2312.16549}.

\bibitem[{Hagstr{\"o}m et~al.(2024)Hagstr{\"o}m, Marjanovi{\'c}, Yu, Arora, Lioma, Maistro, Atanasova, and Augenstein}]{hagstrom2024reality}
Lovisa Hagstr{\"o}m, Sara~Vera Marjanovi{\'c}, Haeun Yu, Arnav Arora, Christina Lioma, Maria Maistro, Pepa Atanasova, and Isabelle Augenstein. 2024.
\newblock A reality check on context utilisation for retrieval-augmented generation.
\newblock \emph{arXiv preprint arXiv:2412.17031}.

\bibitem[{Hou et~al.(2024)Hou, Pascale, Carnerero-Cano, Tchrakian, Marinescu, Daly, Padhi, and Sattigeri}]{hou2024wikicontradict}
Yufang Hou, Alessandra Pascale, Javier Carnerero-Cano, Tigran Tchrakian, Radu Marinescu, Elizabeth Daly, Inkit Padhi, and Prasanna Sattigeri. 2024.
\newblock Wikicontradict: A benchmark for evaluating llms on real-world knowledge conflicts from wikipedia.
\newblock \emph{Advances in Neural Information Processing Systems}, 37:109701--109747.

\bibitem[{Jiang et~al.(2023)Jiang, Sablayrolles, Mensch, Bamford, Chaplot, de~las Casas, Bressand, Lengyel, Lample, Saulnier, Lavaud, Lachaux, Stock, Scao, Lavril, Wang, Lacroix, and Sayed}]{jiang2023mistral7b}
Albert~Q. Jiang, Alexandre Sablayrolles, Arthur Mensch, Chris Bamford, Devendra~Singh Chaplot, Diego de~las Casas, Florian Bressand, Gianna Lengyel, Guillaume Lample, Lucile Saulnier, Lélio~Renard Lavaud, Marie-Anne Lachaux, Pierre Stock, Teven~Le Scao, Thibaut Lavril, Thomas Wang, Timothée Lacroix, and William~El Sayed. 2023.
\newblock \href {http://arxiv.org/abs/2310.06825} {Mistral 7b}.

\bibitem[{Jin et~al.(2024{\natexlab{a}})Jin, Cao, Chen, Liu, Jiang, Xu, Li, and Zhao}]{jin2024tug}
Zhuoran Jin, Pengfei Cao, Yubo Chen, Kang Liu, Xiaojian Jiang, Jiexin Xu, Qiuxia Li, and Jun Zhao. 2024{\natexlab{a}}.
\newblock Tug-of-war between knowledge: Exploring and resolving knowledge conflicts in retrieval-augmented language models.
\newblock \emph{arXiv preprint arXiv:2402.14409}.

\bibitem[{Jin et~al.(2024{\natexlab{b}})Jin, Cao, Yuan, Chen, Xu, Li, Jiang, Liu, and Zhao}]{jin-etal-2024-cutting}
Zhuoran Jin, Pengfei Cao, Hongbang Yuan, Yubo Chen, Jiexin Xu, Huaijun Li, Xiaojian Jiang, Kang Liu, and Jun Zhao. 2024{\natexlab{b}}.
\newblock \href {https://doi.org/10.18653/v1/2024.findings-acl.70} {Cutting off the head ends the conflict: A mechanism for interpreting and mitigating knowledge conflicts in language models}.
\newblock In \emph{Findings of the Association for Computational Linguistics: ACL 2024}, pages 1193--1215, Bangkok, Thailand.

\bibitem[{Kortukov et~al.(2024)Kortukov, Rubinstein, Nguyen, and Oh}]{kortukov2024studying}
Evgenii Kortukov, Alexander Rubinstein, Elisa Nguyen, and Seong~Joon Oh. 2024.
\newblock Studying large language model behaviors under context-memory conflicts with real documents.
\newblock \emph{arXiv preprint arXiv:2404.16032}.

\bibitem[{Landis and Koch(1977)}]{landis1977measurement}
J~Richard Landis and Gary~G Koch. 1977.
\newblock The measurement of observer agreement for categorical data.
\newblock \emph{biometrics}, pages 159--174.

\bibitem[{Lewis et~al.(2020)Lewis, Perez, Piktus, Petroni, Karpukhin, Goyal, K{\"u}ttler, Lewis, Yih, Rockt{\"a}schel et~al.}]{lewis2020retrieval}
Patrick Lewis, Ethan Perez, Aleksandra Piktus, Fabio Petroni, Vladimir Karpukhin, Naman Goyal, Heinrich K{\"u}ttler, Mike Lewis, Wen-tau Yih, Tim Rockt{\"a}schel, et~al. 2020.
\newblock Retrieval-augmented generation for knowledge-intensive nlp tasks.
\newblock \emph{Advances in neural information processing systems}, 33:9459--9474.

\bibitem[{Li et~al.(2025)Li, Chen, and Tong}]{li2025taming}
Gaotang Li, Yuzhong Chen, and Hanghang Tong. 2025.
\newblock Taming knowledge conflicts in language models.
\newblock \emph{arXiv preprint arXiv:2503.10996}.

\bibitem[{Liu et~al.(2025)Liu, Wang, Yuan, Huang, Liu, He, and Tu}]{liu-etal-2025-insight}
Xiaoyuan Liu, Wenxuan Wang, Youliang Yuan, Jen-tse Huang, Qiuzhi Liu, Pinjia He, and Zhaopeng Tu. 2025.
\newblock \href {https://doi.org/10.18653/v1/2025.acl-long.872} {Insight over sight: Exploring the vision-knowledge conflicts in multimodal {LLM}s}.
\newblock In \emph{Proceedings of the 63rd Annual Meeting of the Association for Computational Linguistics (Volume 1: Long Papers)}, pages 17825--17846, Vienna, Austria. Association for Computational Linguistics.

\bibitem[{Liu et~al.(2023)Liu, Iter, Xu, Wang, Xu, and Zhu}]{liu-etal-2023-g}
Yang Liu, Dan Iter, Yichong Xu, Shuohang Wang, Ruochen Xu, and Chenguang Zhu. 2023.
\newblock \href {https://doi.org/10.18653/v1/2023.emnlp-main.153} {{G}-eval: {NLG} evaluation using gpt-4 with better human alignment}.
\newblock In \emph{Proceedings of the 2023 Conference on Empirical Methods in Natural Language Processing}, pages 2511--2522, Singapore.

\bibitem[{Longpre et~al.(2021)Longpre, Perisetla, Chen, Ramesh, DuBois, and Singh}]{longpre-etal-2021-entity}
Shayne Longpre, Kartik Perisetla, Anthony Chen, Nikhil Ramesh, Chris DuBois, and Sameer Singh. 2021.
\newblock \href {https://doi.org/10.18653/v1/2021.emnlp-main.565} {Entity-based knowledge conflicts in question answering}.
\newblock In \emph{Proceedings of the 2021 Conference on Empirical Methods in Natural Language Processing}, pages 7052--7063, Online and Punta Cana, Dominican Republic.

\bibitem[{Mallen et~al.(2023)Mallen, Asai, Zhong, Das, Khashabi, and Hajishirzi}]{mallen-etal-2023-trust}
Alex Mallen, Akari Asai, Victor Zhong, Rajarshi Das, Daniel Khashabi, and Hannaneh Hajishirzi. 2023.
\newblock \href {https://doi.org/10.18653/v1/2023.acl-long.546} {When not to trust language models: Investigating effectiveness of parametric and non-parametric memories}.
\newblock In \emph{Proceedings of the 61st Annual Meeting of the Association for Computational Linguistics (Volume 1: Long Papers)}, pages 9802--9822, Toronto, Canada.

\bibitem[{Meng et~al.(2022)Meng, Bau, Andonian, and Belinkov}]{meng2022locating}
Kevin Meng, David Bau, Alex Andonian, and Yonatan Belinkov. 2022.
\newblock Locating and editing factual associations in gpt.
\newblock \emph{Advances in neural information processing systems}, 35:17359--17372.

\bibitem[{OLMo et~al.(2024)OLMo, Walsh, Soldaini, Groeneveld, Lo, Arora, Bhagia, Gu, Huang, Jordan et~al.}]{olmo20242}
Team OLMo, Pete Walsh, Luca Soldaini, Dirk Groeneveld, Kyle Lo, Shane Arora, Akshita Bhagia, Yuling Gu, Shengyi Huang, Matt Jordan, et~al. 2024.
\newblock 2 olmo 2 furious.
\newblock \emph{arXiv preprint arXiv:2501.00656}.

\bibitem[{OpenAI(2024)}]{openai2024gpt4o}
OpenAI. 2024.
\newblock \href {https://openai.com/index/gpt-4o} {Gpt-4o: Openai’s new flagship model}.
\newblock Accessed: 2025-05-19.

\bibitem[{Perez et~al.(2023)Perez, Ringer, Lukosiute, Nguyen, Chen, Heiner, Pettit, Olsson, Kundu, Kadavath, Jones, Chen, Mann, Israel, Seethor, McKinnon, Olah, Yan, Amodei, Amodei, Drain, Li, Tran-Johnson, Khundadze, Kernion, Landis, Kerr, Mueller, Hyun, Landau, Ndousse, Goldberg, Lovitt, Lucas, Sellitto, Zhang, Kingsland, Elhage, Joseph, Mercado, DasSarma, Rausch, Larson, McCandlish, Johnston, Kravec, El~Showk, Lanham, Telleen-Lawton, Brown, Henighan, Hume, Bai, Hatfield-Dodds, Clark, Bowman, Askell, Grosse, Hernandez, Ganguli, Hubinger, Schiefer, and Kaplan}]{perez-etal-2023-discovering}
Ethan Perez, Sam Ringer, Kamile Lukosiute, Karina Nguyen, Edwin Chen, Scott Heiner, Craig Pettit, Catherine Olsson, Sandipan Kundu, Saurav Kadavath, Andy Jones, Anna Chen, Benjamin Mann, Brian Israel, Bryan Seethor, Cameron McKinnon, Christopher Olah, Da~Yan, Daniela Amodei, Dario Amodei, Dawn Drain, Dustin Li, Eli Tran-Johnson, Guro Khundadze, Jackson Kernion, James Landis, Jamie Kerr, Jared Mueller, Jeeyoon Hyun, Joshua Landau, Kamal Ndousse, Landon Goldberg, Liane Lovitt, Martin Lucas, Michael Sellitto, Miranda Zhang, Neerav Kingsland, Nelson Elhage, Nicholas Joseph, Noemi Mercado, Nova DasSarma, Oliver Rausch, Robin Larson, Sam McCandlish, Scott Johnston, Shauna Kravec, Sheer El~Showk, Tamera Lanham, Timothy Telleen-Lawton, Tom Brown, Tom Henighan, Tristan Hume, Yuntao Bai, Zac Hatfield-Dodds, Jack Clark, Samuel~R. Bowman, Amanda Askell, Roger Grosse, Danny Hernandez, Deep Ganguli, Evan Hubinger, Nicholas Schiefer, and Jared Kaplan. 2023.
\newblock \href {https://doi.org/10.18653/v1/2023.findings-acl.847} {Discovering language model behaviors with model-written evaluations}.
\newblock In \emph{Findings of the Association for Computational Linguistics: ACL 2023}, pages 13387--13434, Toronto, Canada. Association for Computational Linguistics.

\bibitem[{Perez and Ribeiro(2022)}]{perez2022ignore}
F{\'a}bio Perez and Ian Ribeiro. 2022.
\newblock Ignore previous prompt: Attack techniques for language models.
\newblock \emph{arXiv preprint arXiv:2211.09527}.

\bibitem[{Posner and Strike(1992)}]{posner1992revisionist}
George~J Posner and Kenneth~A Strike. 1992.
\newblock A revisionist theory of conceptual change.
\newblock \emph{Philosophy of science, cognitive psychology, and educational theory and practice}, 147.

\bibitem[{Posner et~al.(1982)Posner, Strike, Hewson, Gertzog et~al.}]{posner1982accommodation}
George~J Posner, Kenneth~A Strike, Peter~W Hewson, William~A Gertzog, et~al. 1982.
\newblock Accommodation of a scientific conception: Toward a theory of conceptual change.
\newblock \emph{Science education}, 66(2):211--227.

\bibitem[{Qwen et~al.(2025)Qwen, :, Yang, Yang, Zhang, Hui, Zheng, Yu, Li, Liu, Huang, Wei, Lin, Yang, Tu, Zhang, Yang, Yang, Zhou, Lin, Dang, Lu, Bao, Yang, Yu, Li, Xue, Zhang, Zhu, Men, Lin, Li, Tang, Xia, Ren, Ren, Fan, Su, Zhang, Wan, Liu, Cui, Zhang, and Qiu}]{qwen2025qwen25technicalreport}
Qwen, :, An~Yang, Baosong Yang, Beichen Zhang, Binyuan Hui, Bo~Zheng, Bowen Yu, Chengyuan Li, Dayiheng Liu, Fei Huang, Haoran Wei, Huan Lin, Jian Yang, Jianhong Tu, Jianwei Zhang, Jianxin Yang, Jiaxi Yang, Jingren Zhou, Junyang Lin, Kai Dang, Keming Lu, Keqin Bao, Kexin Yang, Le~Yu, Mei Li, Mingfeng Xue, Pei Zhang, Qin Zhu, Rui Men, Runji Lin, Tianhao Li, Tianyi Tang, Tingyu Xia, Xingzhang Ren, Xuancheng Ren, Yang Fan, Yang Su, Yichang Zhang, Yu~Wan, Yuqiong Liu, Zeyu Cui, Zhenru Zhang, and Zihan Qiu. 2025.
\newblock \href {http://arxiv.org/abs/2412.15115} {Qwen2.5 technical report}.

\bibitem[{Ru et~al.(2024)Ru, Qiu, Hu, Zhang, Shi, Chang, Jiayang, Wang, Sun, Li et~al.}]{ru2024ragchecker}
Dongyu Ru, Lin Qiu, Xiangkun Hu, Tianhang Zhang, Peng Shi, Shuaichen Chang, Cheng Jiayang, Cunxiang Wang, Shichao Sun, Huanyu Li, et~al. 2024.
\newblock Ragchecker: A fine-grained framework for diagnosing retrieval-augmented generation.
\newblock \emph{Advances in Neural Information Processing Systems}, 37:21999--22027.

\bibitem[{Sharma et~al.(2023)Sharma, Tong, Korbak, Duvenaud, Askell, Bowman, Cheng, Durmus, Hatfield-Dodds, Johnston et~al.}]{sharma2024towards}
Mrinank Sharma, Meg Tong, Tomasz Korbak, David Duvenaud, Amanda Askell, Samuel~R Bowman, Newton Cheng, Esin Durmus, Zac Hatfield-Dodds, Scott~R Johnston, et~al. 2023.
\newblock Towards understanding sycophancy in language models.
\newblock \emph{arXiv preprint arXiv:2310.13548}.

\bibitem[{Shi et~al.(2024)Shi, Han, Lewis, Tsvetkov, Zettlemoyer, and Yih}]{shi-etal-2024-trusting}
Weijia Shi, Xiaochuang Han, Mike Lewis, Yulia Tsvetkov, Luke Zettlemoyer, and Wen-tau Yih. 2024.
\newblock \href {https://doi.org/10.18653/v1/2024.naacl-short.69} {Trusting your evidence: Hallucinate less with context-aware decoding}.
\newblock In \emph{Proceedings of the 2024 Conference of the North American Chapter of the Association for Computational Linguistics: Human Language Technologies (Volume 2: Short Papers)}, pages 783--791, Mexico City, Mexico.

\bibitem[{Vosniadou and Brewer(1992)}]{vosniadou1992mental}
Stella Vosniadou and William~F Brewer. 1992.
\newblock Mental models of the earth: A study of conceptual change in childhood.
\newblock \emph{Cognitive psychology}, 24(4):535--585.

\bibitem[{Wan et~al.(2024)Wan, Wallace, and Klein}]{wan-etal-2024-evidence}
Alexander Wan, Eric Wallace, and Dan Klein. 2024.
\newblock \href {https://doi.org/10.18653/v1/2024.acl-long.403} {What evidence do language models find convincing?}
\newblock In \emph{Proceedings of the 62nd Annual Meeting of the Association for Computational Linguistics (Volume 1: Long Papers)}, pages 7468--7484, Bangkok, Thailand.

\bibitem[{Wang et~al.(2025{\natexlab{a}})Wang, Prasad, Stengel-Eskin, and Bansal}]{wang-etal-2025-adacad}
Han Wang, Archiki Prasad, Elias Stengel-Eskin, and Mohit Bansal. 2025{\natexlab{a}}.
\newblock \href {https://aclanthology.org/2025.naacl-long.581/} {{A}da{CAD}: Adaptively decoding to balance conflicts between contextual and parametric knowledge}.
\newblock In \emph{Proceedings of the 2025 Conference of the Nations of the Americas Chapter of the Association for Computational Linguistics: Human Language Technologies (Volume 1: Long Papers)}, pages 11636--11652, Albuquerque, New Mexico. Association for Computational Linguistics.

\bibitem[{Wang et~al.(2022)Wang, Wei, Schuurmans, Le, Chi, and Zhou}]{wang2022rationale}
Xuezhi Wang, Jason Wei, Dale Schuurmans, Quoc Le, Ed~Chi, and Denny Zhou. 2022.
\newblock Rationale-augmented ensembles in language models.
\newblock \emph{arXiv preprint arXiv:2207.00747}.

\bibitem[{Wang et~al.(2025{\natexlab{b}})Wang, Wang, Bai, and Luo}]{wang-etal-2025-continuously}
Yilin Wang, Heng Wang, Yuyang Bai, and Minnan Luo. 2025{\natexlab{b}}.
\newblock \href {https://doi.org/10.18653/v1/2025.emnlp-main.233} {Continuously steering {LLM}s sensitivity to contextual knowledge with proxy models}.
\newblock In \emph{Proceedings of the 2025 Conference on Empirical Methods in Natural Language Processing}, pages 4682--4698, Suzhou, China. Association for Computational Linguistics.

\bibitem[{Wei et~al.(2023)Wei, Huang, Lu, Zhou, and Le}]{wei2024simple}
Jerry Wei, Da~Huang, Yifeng Lu, Denny Zhou, and Quoc~V Le. 2023.
\newblock Simple synthetic data reduces sycophancy in large language models.
\newblock \emph{arXiv preprint arXiv:2308.03958}.

\bibitem[{Wu et~al.(2024)Wu, Zhang, and Zhao}]{wu2024mitigating}
Yexin Wu, Zhuosheng Zhang, and Hai Zhao. 2024.
\newblock Mitigating misleading chain-of-thought reasoning with selective filtering.
\newblock \emph{arXiv preprint arXiv:2403.19167}.

\bibitem[{Xie et~al.(2023)Xie, Zhang, Chen, Lou, and Su}]{xie2023adaptive}
Jian Xie, Kai Zhang, Jiangjie Chen, Renze Lou, and Yu~Su. 2023.
\newblock Adaptive chameleon or stubborn sloth: Revealing the behavior of large language models in knowledge conflicts.
\newblock In \emph{The Twelfth International Conference on Learning Representations}.

\bibitem[{Xu et~al.(2024)Xu, Qi, Guo, Wang, Wang, Zhang, and Xu}]{xu-etal-2024-knowledge-conflicts}
Rongwu Xu, Zehan Qi, Zhijiang Guo, Cunxiang Wang, Hongru Wang, Yue Zhang, and Wei Xu. 2024.
\newblock \href {https://doi.org/10.18653/v1/2024.emnlp-main.486} {Knowledge conflicts for {LLM}s: A survey}.
\newblock In \emph{Proceedings of the 2024 Conference on Empirical Methods in Natural Language Processing}, pages 8541--8565, Miami, Florida, USA.

\bibitem[{Ying et~al.(2024)Ying, Cao, Xiong, Cui, He, and Liu}]{ying2024intuitive}
Jiahao Ying, Yixin Cao, Kai Xiong, Long Cui, Yidong He, and Yongbin Liu. 2024.
\newblock \href {https://doi.org/10.18653/v1/2024.acl-long.232} {Intuitive or dependent? investigating {LLM}s' behavior style to conflicting prompts}.
\newblock In \emph{Proceedings of the 62nd Annual Meeting of the Association for Computational Linguistics (Volume 1: Long Papers)}, pages 4221--4246, Bangkok, Thailand. Association for Computational Linguistics.

\bibitem[{Zagzebski(1999)}]{Zagzebski1999-ZAGWIK}
Linda Zagzebski. 1999.
\newblock "what is knowledge?".
\newblock In John Greco and Ernest Sosa, editors, \emph{The Blackwell Guide to Epistemology}, pages 92--116. Wiley-Blackwell.

\bibitem[{Zheng et~al.(2023)Zheng, Chiang, Sheng, Zhuang, Wu, Zhuang, Lin, Li, Li, Xing et~al.}]{zheng2023judging}
Lianmin Zheng, Wei-Lin Chiang, Ying Sheng, Siyuan Zhuang, Zhanghao Wu, Yonghao Zhuang, Zi~Lin, Zhuohan Li, Dacheng Li, Eric Xing, et~al. 2023.
\newblock Judging llm-as-a-judge with mt-bench and chatbot arena.
\newblock \emph{Advances in Neural Information Processing Systems}, 36:46595--46623.

\end{thebibliography}
\bibliographystyle{acl_natbib}

\appendix

\section{Parametric Knowledge Query}
\label{app:parametricprompt}
We query for the parametric knowledge with multiple prompts.
For a single instance $(q_i, \{a_{i1}, a_{i2}\}, \{c_{i1}, c_{i2}\} )$ in dataset $D_{\text{orig}}=\{(q_i, \{a_{i1}, a_{i2}\}, \{c_{i1}, c_{i2}\} ), i\in[1, N]\}$, we prompt the model to confirm whether they believe the answer to $q_i$ is $a_{i1}$ or $a_{i2}$.
If the model deems one of the $a_{ij}$'s as the only correct answer to question $q_i$, this instance will be included in the parametric knowledge base, and $a_{ij}$ will be assigned as No Contradiction (\texttt{NC}) passage.
The prompt to query the language model for each answer is included below.
\vspace{20pt}
\fbox{%
  \parbox{\dimexpr\linewidth-2\fboxsep-2\fboxrule}{%
    \ttfamily
    You are an independent model with rich knowledge. You will be asked to validate whether the given answer is correct, and you should solely give your judgment in the form of yes or no without additional information.\\
    Question: \{question\}\\
    Answer: \{answer\}\\
    Is this answer correct? <think>
  }%
}

\section{Verification of Data Validity}
\label{app:validation}
In the final stage of the diagnostic data creation flow (\texttt{Validation} in \Cref{fig:dataflow}), all instances are verified by two language models and subsampled for human verification.
The number of instances removed during this process is reported in \Cref{tab:instance_removal}.

\paragraph{Model Verification} 
Each (question, evidence, answer) triple was validated by GPT-4o and DeepSeek-R1-Distill-Llama-70B \cite{guo2025deepseek}. 
Both models were required to answer the question using the provided gold evidence. 
An instance was retained only if both predictions matched the annotated answer.

\paragraph{Human Verification} 
To further ensure data quality, a subset of model-verified instances was manually reviewed.
For each test model (OLMo, Mistral, and Qwen), 30 randomly selected instances were manually reviewed. Annotators verified (1) whether the evidence supported its answer, and (2) whether the gold answer was correct for the corresponding task (e.g., CK answer for CK task; PK answer for PK task). 
All sampled instances passed human verification, likely reflecting the strictness of the preceding model-based filter.
The annotators were graduate students in NLP with prior annotation experience. 
They were compensated at standard research assistant rates.

\section{Dataset Statistics}
\label{app:stat}
Because each model encodes different parametric knowledge, the resulting diagnostic datasets differ across models.
The overall statistics are reported in \Cref{tab:diagnostic_data}. 
Each question is paired with four evidence conditions (\texttt{NC}, \texttt{HPC}, \texttt{HPCE}, \texttt{LPC}), so the effective dataset size is four times the number of base instances.

\begin{table}[H]
\centering
\begin{tabular}{l r}
\toprule
\textbf{Model} & \textbf{\# Instances} \\
\midrule
Mistral-7B   & 2,893 \\
OLMo2-7B     &   177 \\
OLMo2-13B    &   456 \\
Qwen2.5-7B   & 6,217 \\
Qwen2.5-14B  & 4,250 \\
GPT-5.2      & 7,250 \\
\bottomrule
\end{tabular}
\caption{Number of instances of the resulting data for each model.}
\label{tab:diagnostic_data}
\end{table}

\begin{table*}[t]
\centering
\small
\begin{tabular}{lcccccc}
\toprule
\textbf{Stage} & \textbf{Mistral} & \textbf{OLMo2-7B} & \textbf{OLMo2-13B} & \textbf{Qwen2.5-7B} & \textbf{Qwen2.5-14B} & \textbf{GPT-5.2} \\
\midrule
Parametric knowledge querying (\# dropped) & 10,752 & 13,487 & 13,207 & 7,432 & 9,379 & -- \\
Model verification (\# dropped questions)  & 22     & 4      & 4      & 18    & 38    & -- \\
Model verification (\# dropped instances)  & 88     & 16     & 16     & 72    & 152   & -- \\
Final (\# questions)                        & 2,893  & 177    & 456    & 6,217 & 4,250 & 7,250 \\
\bottomrule
\end{tabular}
\caption{Instances removed during each stage of data creation and filtering. Each question is associated with four evidence types, so the total number of evidence–answer pairs equals four times the final question count; the number of dropped instances scales accordingly. Per-stage drop counts for GPT-5.2 are not reported as the same parametric-knowledge filter and model-verification stages were rerun against GPT-5.2 only at the final-instance level.}
\label{tab:instance_removal}
\end{table*}

\section{Task Examples}
\label{app:sec:taskexample}
\begin{figure*}[h]           %
  \centering
  \begin{subfigure}[h]{\the\textwidth}
    \centering
  \begin{fullwidthExampleBox}
  [title={Knowledge Free Task Example}]
    \begin{tabularx}{\textwidth}{@{}lX@{}}  %
      \textbf{Input} & You are an extractive question-answering model. Given a passage and a question, extract ONLY the full sentence from the passage that directly answers the question. Do not generate summaries or paraphrase. Only return the complete sentence that contains the answer. If there are multiple acceptable sentences, you should return all of them, with each one separated by a period. Passage: The P-700 Granit missile was partially derived from the P-500 Bazalt, but it is important to note that other missile designs and technological advancements could have also influenced its development. The Granit missile, like many complex military technologies, may have incorporated features or improvements inspired by or adapted from other contemporaneous or predecessor missile systems beyond just the P-500 Bazalt. Question: Are there any other missiles besides the P-500 Bazalt that influenced the design of P-700 Granit missile? Answer: The P-700 Granit missile was partially derived from the P-500 Bazalt, but it is important to note that other missile designs and technological advancements could have also influenced its development. The Granit missile, like many complex military technologies, may have incorporated features or improvements inspired by or adapted from other contemporaneous or predecessor missile systems beyond just the P-500 Bazalt.
      Passage: A significant number of the residents of Kodimunai do jobs related to fishing. These jobs includes deep sea fishing, shallow water fishing, fishing from the shore (known as karamadi in the local language), fishing with mechanized boats, exporting fish, etc.
      Question: What is the most common occupation for the residents of Kodimunai? Answer:  \\
      \textbf{Gold Answer}   & A significant number of the residents of Kodimunai do jobs related to fishing \\[4pt]
    \end{tabularx}
  \end{fullwidthExampleBox}
  \caption{Example of knowledge-free task. The model is expected to extract the answer directly from the context.}
  \label{fig:kf-example}
  \end{subfigure}
    \hfill
  \begin{subfigure}[h]{\the\textwidth}
    \centering
  \begin{fullwidthExampleBox}
  [title={Contextual Knowledge Task Example}]
    \begin{tabularx}{\textwidth}{@{}lX@{}}  %
      \textbf{Input} & You are a question-answering system that strictly answers questions based only on the given context. Do not use external knowledge or make assumptions beyond what is explicitly stated. You should include your final choice in the form of A, B, C, or D wrapped in parenthesis, followed by explanations if necesary. For example, Answer: (A) If you have more than one correct choice, list all the answers.
      Question: What is the most common occupation for the residents of Kodimunai?
      Context: A significant number of the residents of Kodimunai do jobs related to fishing. These jobs includes deep sea fishing, shallow water fishing, fishing from the shore (known as karamadi in the local language), fishing with mechanized boats, exporting fish, etc.
      Choices: A.Aerospace engineering B.Fishing C.IT, medicine, engineering, trading D.Farming 
      Answer: \\
      \textbf{Gold Answer}   & B \\[4pt]
    \end{tabularx}
  \end{fullwidthExampleBox}
  \caption{Example of contextual knowledge task.}
  \label{fig:ck-example}
  \end{subfigure}
  \hfill
  \begin{subfigure}[h]{\the\textwidth}
    \centering
  \begin{fullwidthExampleBox}
  [title={Parametric Knowledge Task Example}]
    \begin{tabularx}{\textwidth}{@{}lX@{}}  %
      \textbf{Input} & "You are a knowledgeable question-answering system. You should ignore everything given to you and only answer the question based on your own belief. You can provide justification if needed. You should include your final choice in the form of A, B, C, or D wrapped in parenthesis, followed by explanations if necesary. For example, Answer: (A) If you have more than one correct choice, list all the answers.Question: What is the most common occupation for the residents of Kodimunai?
      Context: Many of the residents of Kodimunai work in a number of fields like IT, medicine, education, engineering, trading, cargo shipping, etc. 
      Choices: A.Aerospace engineering B.Fishing C.IT, medicine, engineering, trading D.Farming 
      Answer:  \\
      \textbf{Gold Answer}   & B \\[4pt]
    \end{tabularx}
  \end{fullwidthExampleBox}
  \caption{Example of parametric knowledge task. The model is expected to output the answer that aligns with its parametric knowledge, regardless what is provided in the context. Here, the model's parametric knowledge is B. Fishing.}
  \label{fig:pk-example}
  \end{subfigure}

\caption{Examples of each task.}
  \label{fig:task-examples}
\end{figure*}

\begin{figure*}[h]           %
  \centering
  \begin{subfigure}[h]{\the\textwidth}
    \centering
  \begin{fullwidthExampleBox}
  [title={Parametrick Contextual Task Example}]
    \begin{tabularx}{\textwidth}{@{}lX@{}}  %
      \textbf{Input} & You are a knowledgeable question-answering system. You will be given a context, a question, and a list of choices. Your task is to answer the question using your best possible knowledge. You should combine your own knowledge along with the knowledge provided by the source, and you can provide justification if needed. Note that the provided source is not always reliable. You should include your final choice in the form of A, B, C, or D wrapped in parentheses, followed by explanations if necessary. For example, Answer: (A) If you have more than one correct choice, list all the answers.Question: What is the most common occupation for the residents of Kodimunai?
      Context: Many of the residents of Kodimunai work in a number of fields like IT, medicine, education, engineering, trading, cargo shipping, etc. 
      Choices: A.Aerospace engineering B.Fishing C.IT, medicine, engineering, trading D.Farming
      Answer:   \\
      \textbf{Gold Answer}   & CB \\[4pt]
    \end{tabularx}
  \end{fullwidthExampleBox}
  \caption{Example of \texttt{PCK} task. The model is given only an external context and is expected to combine its parametric knowledge along with the external knowledge to provide the answer.}
  \label{fig:pck-example}
  \end{subfigure}
    \hfill
  \begin{subfigure}[h]{\the\textwidth}
    \centering
  \begin{fullwidthExampleBox}
  [title={Retrieval Augmented Generation Task Example}]
    \begin{tabularx}{\textwidth}{@{}lX@{}}  %
      \textbf{Input} & Select the correct answers for the following question based on the given contexts. Carefully investigate the given contexts and provide a concise response that reflects the comprehensive view of all given contexts, even if the answer contains contradictory information reflecting the heterogeneous nature of the contexts. You should include your final choice in the form of A, B, C, or D wrapped in parentheses, followed by explanations if necessary. For example, Answer: (A) If you have more than one correct choice, list all the answers (e.g., Answer: (BC)).
      Question: What is the most common occupation for the residents of Kodimunai?
      Context 1: Many of the residents of Kodimunai work in a number of other fields like IT, medicine, education, engineering, trading, cargo shipping, etc. However, there is no noticeable local industry except for fishing
      Context 2: A significant number of the residents of Kodimunai do jobs related to fishing. These jobs include deep sea fishing, shallow water fishing, fishing from the shore (known as karamadi in the local language), fishing with mechanized boats, exporting fish, etc.
      Choices: A.Aerospace engineering B.Fishing C.IT, medicine, engineering, trading D.Farming 
      Answer:  \\
      \textbf{Gold Answer}   & BC \\[4pt]
    \end{tabularx}
  \end{fullwidthExampleBox}
  \caption{Example of \texttt{RAG} task. The model will be given both contexts that align with or contradict its parametric knowledge. It is expected to provide the answer based on both contexts.}
  \label{fig:rag-example}
  \end{subfigure}

\caption{Examples of each task.(cont)}
  \label{fig:task-examples2}
\end{figure*}

Examples of each task are provided in \Cref{fig:task-examples} and \Cref{fig:task-examples2}.

\section{Raw Performance}
\label{app:sec:rawperfnumber}
\begin{table*}[t]
\centering
\small
\setlength{\tabcolsep}{6pt}
\begin{tabular}{l l r r r r}
\toprule
\textbf{Model} & \textbf{Task} & \textbf{NC} & \textbf{HPC} & \textbf{HPCE} & \textbf{LPC}\\
\midrule
\multirow{10}{*}{Mistral-7B} & \multirow{2}{*}{KFextract} & 1.6 & 1.4 & 0.5 & 0.4\\
 &  & \cellcolor{gray!15} 76.7 & \cellcolor{gray!15} 77.3 & \cellcolor{gray!15} 83.4 & \cellcolor{gray!15} 85.2\\
 & \multirow{2}{*}{CK} & 65.3 & 28.5 & 42.2 & 38.2\\
 &  & \cellcolor{gray!15} 65.3 & \cellcolor{gray!15} 28.9 & \cellcolor{gray!15} 42.3 & \cellcolor{gray!15} 38.5\\
 & \multirow{2}{*}{PK} & 62.6 & 18.4 & 6.4 & 51.4\\
 &  & \cellcolor{gray!15} 62.7 & \cellcolor{gray!15} 18.6 & \cellcolor{gray!15} 6.5 & \cellcolor{gray!15} 51.6\\
 & \multirow{2}{*}{PCK} & 62.4 & 0 & 0 & 9.4\\
 &  & \cellcolor{gray!15} 62.5 & \cellcolor{gray!15} 31.5 & \cellcolor{gray!15} 34.2 & \cellcolor{gray!15} 54.5\\
 & \multirow{2}{*}{RAG} & 54.4 & 0.2 & 0.1 & 7.9\\
 &  & \cellcolor{gray!15} 54.7 & \cellcolor{gray!15} 41.2 & \cellcolor{gray!15} 38.1 & \cellcolor{gray!15} 54.4\\
\midrule
\multirow{10}{*}{OLMo2-7B} & \multirow{2}{*}{KFextract} & 0 & 0 & 0.7 & 0\\
 &  & \cellcolor{gray!15} 54.4 & \cellcolor{gray!15} 66.2 & \cellcolor{gray!15} 66.4 & \cellcolor{gray!15} 74.3\\
 & \multirow{2}{*}{CK} & 56.8 & 48.9 & 47.2 & 55.1\\
 &  & \cellcolor{gray!15} 57 & \cellcolor{gray!15} 48.9 & \cellcolor{gray!15} 48.5 & \cellcolor{gray!15} 56\\
 & \multirow{2}{*}{PK} & 55.7 & 11.9 & 7.4 & 31.2\\
 &  & \cellcolor{gray!15} 58.5 & \cellcolor{gray!15} 14.3 & \cellcolor{gray!15} 9 & \cellcolor{gray!15} 32.2\\
 & \multirow{2}{*}{PCK} & 44.3 & 0 & 0 & 5.7\\
 &  & \cellcolor{gray!15} 48.2 & \cellcolor{gray!15} 39 & \cellcolor{gray!15} 40.8 & \cellcolor{gray!15} 50.3\\
 & \multirow{2}{*}{RAG} & 41.5 & 1.1 & 0 & 3.4\\
 &  & \cellcolor{gray!15} 47.8 & \cellcolor{gray!15} 43.3 & \cellcolor{gray!15} 42.4 & \cellcolor{gray!15} 46.3\\
\midrule
\multirow{10}{*}{OLMo2-13B} & \multirow{2}{*}{KFextract} & 1.9 & 0.2 & 0.2 & 0.7\\
 &  & \cellcolor{gray!15} 67.4 & \cellcolor{gray!15} 67.3 & \cellcolor{gray!15} 74.7 & \cellcolor{gray!15} 73.4\\
 & \multirow{2}{*}{CK} & 25.9 & 16 & 20.2 & 4.8\\
 &  & \cellcolor{gray!15} 26.4 & \cellcolor{gray!15} 17 & \cellcolor{gray!15} 21.3 & \cellcolor{gray!15} 6\\
 & \multirow{2}{*}{PK} & 57.5 & 31.6 & 18 & 50.4\\
 &  & \cellcolor{gray!15} 57.6 & \cellcolor{gray!15} 32 & \cellcolor{gray!15} 18 & \cellcolor{gray!15} 50.4\\
 & \multirow{2}{*}{PCK} & 76.1 & 0 & 0 & 6.6\\
 &  & \cellcolor{gray!15} 76.1 & \cellcolor{gray!15} 57.2 & \cellcolor{gray!15} 50.4 & \cellcolor{gray!15} 42.7\\
 & \multirow{2}{*}{RAG} & 82 & 0.4 & 0 & 12.7\\
 &  & \cellcolor{gray!15} 82.4 & \cellcolor{gray!15} 60.9 & \cellcolor{gray!15} 60.5 & \cellcolor{gray!15} 65.8\\
\midrule
\multirow{10}{*}{Qwen2.5-7B} & \multirow{2}{*}{KFextract} & 1.6 & 0.9 & 0.3 & 0.5\\
 &  & \cellcolor{gray!15} 75.1 & \cellcolor{gray!15} 71.4 & \cellcolor{gray!15} 78 & \cellcolor{gray!15} 80.4\\
 & \multirow{2}{*}{CK} & 78.8 & 47.2 & 57.4 & 42.6\\
 &  & \cellcolor{gray!15} 82.1 & \cellcolor{gray!15} 50.3 & \cellcolor{gray!15} 64.4 & \cellcolor{gray!15} 48.2\\
 & \multirow{2}{*}{PK} & 82.8 & 48.5 & 28.1 & 62.5\\
 &  & \cellcolor{gray!15} 84.4 & \cellcolor{gray!15} 50.3 & \cellcolor{gray!15} 31.2 & \cellcolor{gray!15} 65.8\\
 & \multirow{2}{*}{PCK} & 83.9 & 0.5 & 0.9 & 14.3\\
 &  & \cellcolor{gray!15} 84.4 & \cellcolor{gray!15} 61.5 & \cellcolor{gray!15} 60.5 & \cellcolor{gray!15} 69\\
 & \multirow{2}{*}{RAG} & 79.5 & 1.3 & 1.5 & 14.3\\
 &  & \cellcolor{gray!15} 82.3 & \cellcolor{gray!15} 60.2 & \cellcolor{gray!15} 59.8 & \cellcolor{gray!15} 67.1\\
\midrule
\multirow{10}{*}{Qwen2.5-14B} & \multirow{2}{*}{KFextract} & 1 & 0.7 & 0.8 & 0.4\\
 &  & \cellcolor{gray!15} 78.4 & \cellcolor{gray!15} 79.5 & \cellcolor{gray!15} 82.8 & \cellcolor{gray!15} 81.2\\
 & \multirow{2}{*}{CK} & 87.4 & 70.2 & 85.5 & 41.6\\
 &  & \cellcolor{gray!15} 90.5 & \cellcolor{gray!15} 73.2 & \cellcolor{gray!15} 90.5 & \cellcolor{gray!15} 44\\
 & \multirow{2}{*}{PK} & 76.7 & 25.7 & 4.1 & 48.4\\
 &  & \cellcolor{gray!15} 86.9 & \cellcolor{gray!15} 32 & \cellcolor{gray!15} 9 & \cellcolor{gray!15} 55.1\\
 & \multirow{2}{*}{PCK} & 84.3 & 2.1 & 1.8 & 6.8\\
 &  & \cellcolor{gray!15} 89.9 & \cellcolor{gray!15} 62.7 & \cellcolor{gray!15} 63.7 & \cellcolor{gray!15} 61.1\\
 & \multirow{2}{*}{RAG} & 82.4 & 10.2 & 9.9 & 9.7\\
 &  & \cellcolor{gray!15} 87.9 & \cellcolor{gray!15} 59.1 & \cellcolor{gray!15} 59.9 & \cellcolor{gray!15} 61\\
\bottomrule
\end{tabular}
\caption{Performance of models. Exact Match rows are unshaded; F1 rows are shaded in numeric columns.}
\label{tab:model_performance}
\end{table*}

We measure both F1 and exact match of each setting.
The full performance of each model on the diagnostic data is shown in \Cref{tab:model_performance}.

\subsection{Highly Confident Instances}
\begin{table}[H]
\centering
\small
\begin{tabular}{@{}clrrrr@{}}
\toprule
\multicolumn{1}{l}{\textbf{Model}} & \textbf{Task} & \multicolumn{1}{l}{\textbf{NC}} & \multicolumn{1}{l}{\textbf{HPC}} & \multicolumn{1}{l}{\textbf{HPCE}} & \multicolumn{1}{l}{\textbf{LPC}} \\ \midrule
\multirow{4}{*}{Mistral-7B} & CK & 100 & 62.8 & 57.2 & 51.4 \\
 & PK & 100 & 63.5 & 43.7 & 45.3 \\
 & PCK & 100 & 50.0 & 33.3 & 27.7 \\
 & RAG & 100 & 50.8 & 33.8 & 28.5 \\
 \midrule
\multirow{4}{*}{OLMo2-7B} & CK & 100 & 87.5 & 79.2 & 78.1 \\
 & PK & 100 & 50.0 & 33.3 & 25.0 \\
 & PCK & 100 & 50.0 & 33.3 & 25.0 \\
 & RAG & 100 & 50.0 & 33.3 & 25.0 \\
 \midrule
\multirow{4}{*}{Qwen2.5-7B} & CK & 100 & 71.4 & 66.3 & 61.6 \\
 & PK & 100 & 75.6 & 59.0 & 59.2 \\
 & PCK & 100 & 50.9 & 34.1 & 28.9 \\
 & RAG & 100 & 51.6 & 34.8 & 29.9 \\ 
 \bottomrule
\end{tabular}
\caption{Performance of models on highly confident instances.}
\label{app:tab:highconf_perf}
\end{table}

\label{app:sec:rawperfnumber:conf}
When querying for the model's parametric knowledge (\texttt{parametric knowledge collection} in \cref{fig:dataflow}), model responses to queries are collected in a binary stance format (e.g., yes/no).
However, when prompted with free-form generation followed by multiple-choice selection, models do not always achieve perfect accuracy on \texttt{NC} instances (\cref{fig:perf}). 
To isolate this effect, we select only the instances that models answer with 100\% accuracy in the \texttt{NC} condition, thereby restricting analysis to fully mastered samples.
The performance of each model on only the highly confident instances is included in \Cref{app:tab:highconf_perf}.
The results confirm that while the absolute numbers vary slightly, the overall trends observed in the broader dataset persist.

\section{Instruction Strength}
\label{app:instruction-strength}
To further disentangle the role of instruction-following ability from conflict-resolution biases, we conducted an additional experiment varying the strength of instructions with Mistral-7B and OLMo-7B. Specifically, we applied three levels of instruction forcefulness:

\begin{itemize}
    \item \textbf{Strong:} ``You \textbf{MUST} strictly and exclusively use~$\ldots$''
    \item \textbf{Neutral:} ``Answer the question based only on~$\ldots$''
    \item \textbf{Weak:} ``Try to answer based on~$\ldots$''
\end{itemize}
The average performance on each strength level is shown in \Cref{app:tab:strengh-avg-perf}, and the fine-grained performance on each context and task type is shown in \Cref{app:tab:mistralstrength}
Instruction strength influences absolute performance, but the impact varies across models. Mistral-7B remains relatively stable under different prompt formulations, whereas OLMo-7B shows sharp degradation under strong, restrictive instructions.
Despite shifts in absolute scores, the relative patterns reported in \sect{sec:findings} remain consistent.
Performance differences across tasks and evidence types are preserved, and the ordering of task/evidence effects holds across weak, neutral, and strong prompts. 

\begin{table}[H]
\small
\centering
\begin{tabular}{lccc}
\toprule
\textbf{Model} & Weak & Neutral & Strong \\
\midrule
Mistral-7B & 45.66 & 46.49 & 49.48 \\
OLMo2-7B   & 45.21 & 44.52 & 30.83 \\
\bottomrule
\end{tabular}
\caption{Average performance on each prompt strength.}
\label{app:tab:strengh-avg-perf}
\end{table}

\begin{table}[H]
\centering
\small
\begin{tabular}{llcccc}
\toprule
\textbf{Task} & \textbf{Strength} & NC & HPC & HPCE & LPC \\
\midrule
\multirow{3}{*}{CK} 
  & Weak    & 43.69 & 57.79 & 69.34 & 53.16 \\
  & Neutral & 44.76 & 57.52 & 71.52 & 57.90 \\
  & Strong  & 46.39 & 53.85 & 70.03 & 54.20 \\
\midrule
\multirow{3}{*}{PK} 
  & Weak    & 47.91 & 18.53 &  7.95 & 36.50 \\
  & Neutral & 46.60 & 15.31 &  7.60 & 37.47 \\
  & Strong  & 55.06 & 23.06 &  8.78 & 46.98 \\
\midrule
\multirow{3}{*}{PCK} 
  & Weak    & 44.88 & 46.45 & 52.64 & 57.95 \\
  & Neutral & 41.38 & 50.58 & 53.52 & 58.28 \\
  & Strong  & 49.59 & 48.13 & 53.59 & 55.78 \\
\midrule
\multirow{3}{*}{RAG} 
  & Weak    & 51.23 & 45.95 & 45.65 & 50.93 \\
  & Neutral & 51.50 & 49.36 & 49.71 & 50.80 \\
  & Strong  & 57.02 & 57.21 & 56.42 & 55.54 \\
\bottomrule
\end{tabular}
\caption{Performance of Mistral-7B with different instruction strength. While absolute performance varies, the relative performance relationship still holds.}
\label{app:tab:mistralstrength}
\end{table}

\section{Explanation Generation for HPCE}
\label{sec:app:expgen}
When encountering context that conflicts with prior knowledge, humans are often more persuaded by additional explanations, which help them iteratively update their mental models with new information \cite{vosniadou1992mental}.
To study this effect, we augment \texttt{HPC} passages with free-text rationales that explicitly reconcile the contradiction from the model-aligned \texttt{NC} perspective.
We denote these augmented passages as \texttt{HPCE} (High Plausibility Contradiction with Explanation).
The explanations are generated by providing both the \texttt{NC} and \texttt{HPC} answers to a language model and prompting it to produce a corresponding rationale.
An example of an \texttt{HPCE} passage is shown in \Cref{fig:hpce-example}, and the full prompt used for explanation generation is provided below.

\vspace{10pt}
\fbox{%
  \parbox{\dimexpr\linewidth-2\fboxsep-2\fboxrule}{%
    \ttfamily
    Based on the given passage, write a coherent and informative passage that naturally explains why \{$a^{\texttt{HPC}}$\} is the correct explanation or conclusion to the question {$q$} instead of \{$a^{\texttt{NC}}$\}. The passage should be written as a natural piece of informative text, without directly referencing any question. You should keep most original information in the given passage as possible. Ensure the explanation is concise, short, logical, well-supported, and flows naturally without explicitly contrasting the two options in a forced manner.
  }%
}

\begin{figure*}[t]           %
  \centering
  \begin{fullwidthExampleBox}
  []
    \begin{tabularx}{\textwidth}{@{}lX@{}}  %
      \textbf{HPC} & Many of the residents of Kodimunai work in a number of other fields like IT, medicine, education, engineering, trading, cargo shipping, etc.  \\
      \textbf{HPCE} &  Despite the historic presence of fishing as a key activity in Kodimunai, the livelihood dynamics in the area have experienced a significant shift towards professional sectors such as IT, medicine, engineering, and trading. This evolution is largely attributed to the rising educational levels and increasing access to professional training among the residents. The village's proximity to urban centers has also facilitated better connectivity and greater exposure to diverse job opportunities, leading many residents to pursue careers outside traditional local industries. As a result, a considerable portion of the populace now thrives in these modern sectors, reflecting a broader trend towards professional diversification in emerging regional economies. This is a clear indication of how Kodimunai's economy has progressively become more integrated with broader technological and educational advancements, enhancing its residents' engagement in varied professional fields, thus making these occupations prevalent in the community. \\[4pt]
    \end{tabularx}
  \end{fullwidthExampleBox}
\caption{An example of \texttt{HPC} and \texttt{HPCE}.}
  \label{fig:hpce-example}
\end{figure*}

\section{Free Generation Setting}
\subsection{Evaluator Prompts}
\label{sec:evalprompt}
\begin{figure*}[t]
  \centering
    \definecolor{partial}{RGB}{255,245,200}   %
    \definecolor{incorrect}{RGB}{255,204,216} %
    \definecolor{correct}{RGB}{187,220,255}   %
\resizebox{\linewidth}{!}{%
\begin{tikzpicture}[
  >=Stealth,
  every edge/.style={draw,->,line width=.8pt},
  decision/.style  ={rectangle, draw, fill=gray!10,
                     text width=5cm, align=center, minimum height=12mm},
  partial/.style   ={rectangle, draw, fill=partial,
                     text width=1.6cm, align=center, minimum height=12mm},
  incorrect/.style ={rectangle, draw, fill=incorrect,
                     text width=1.6cm, align=center, minimum height=12mm},
  correct/.style   ={rectangle, draw, fill=correct,
                     text width=1.5cm, align=center, minimum height=12mm},
]

\node[decision] (root) at (0,0)
      {Contains all correct answers?};

\node[decision] (add) at (-4,-2.6)
      {Contains additional answers not in acceptable list?};
\node[decision] (has) at ( 4,-2.6)
      {Contains at least one correct answer?};

\node[decision] (one1) at (-8,-5)
      {Is there only one gold answer?};
\node[decision] (one2) at (-2,-5)
      {Is there only one gold answer?};

\node[partial]   (p3)   at ( 2.5,-5) {Partially correct};
\node[incorrect] (inc1) at ( 6.5,-5) {Incorrect};

\node[incorrect] (inc0) at (-10,-7.3) {Incorrect};
\node[partial]   (p1)   at ( -7,-7.3) {Partially correct};

\node[correct]   (corr1) at (-4.5,-7.3) {Correct};
\node[decision]  (pref)  at (-0.3,-7.3)
      {The output prefer one answer over another?};

\node[partial]  (p2)    at (-2,-9.6) {Partially correct};
\node[decision] (blend) at ( 3,-9.6)
      {Does the output merely blend the answers without indicating the conflict?};

\node[incorrect] (inc2) at ( 1.2,-11.9) {Incorrect};
\node[correct]   (corr2) at ( 5,-11.9)  {Correct};

\draw (root) -- node[above,sloped] {yes} (add);
\draw (root) -- node[above,sloped] {no}  (has);

\draw (add)  -- node[above,sloped] {yes} (one1);
\draw (add)  -- node[above,sloped] {no}  (one2);

\draw (has)  -- node[above,sloped] {yes} (p3);
\draw (has)  -- node[above,sloped] {no}  (inc1);

\draw (one1) -- node[above,sloped] {yes} (inc0);
\draw (one1) -- node[above,sloped] {no}  (p1);

\draw (one2) -- node[above,sloped] {yes} (corr1);
\draw (one2) -- node[above,sloped] {no}  (pref);

\draw (pref) -- node[above,sloped] {yes} (p2);
\draw (pref) -- node[above,sloped] {no}  (blend);

\draw (blend) -- node[above,sloped] {yes} (inc2);
\draw (blend) -- node[above,sloped] {no}  (corr2);

\end{tikzpicture}%
}
  
  \caption{Decision tree for annotation guidance.}
  \label{app:fig:annotation-tree}
\end{figure*}

In the free generation setting described in \sect{sec:res:evalcase}, we employ a language model as an evaluator to assess the quality of generated answers.  
We experiment with several evaluation prompts and select the one that yields the highest agreement with human annotations.  
Following the design of the evaluator proposed in \citet{hou2024wikicontradict}, we introduce modifications to improve reliability, achieving a higher Cohen’s $\kappa$ with human annotators.  
The final evaluator prompt is provided in \Cref{app:tab:freeevalprompt}, and a decision tree illustrating the evaluation process is shown in \Cref{app:fig:annotation-tree}.

\begin{figure*}[h]  %
  \centering
  \begin{tcolorbox}[           %
    enhanced,
    title={\normalsize Free Generation Evaluation Prompt 1.},
    colback=gray!3,
    colframe=gray!60,
    left=1mm,right=1mm,
    top=0.6em,bottom=0.6em,
    boxrule=0.4pt
  ]
    \begin{lstlisting}
Do not fact-check or rely on world knowledge. Assume the reference (correct_answers) is correct. Please evaluate the response to a question under relaxed evaluation, where hallucinations, outdated information are allowed, as long as the primary answer is accurate. Each response is evaluated as correct, partially correct, or incorrect. The list of acceptable answers is separated by "|". You can discuss your reasoning after "comment: ", but you should include your final judgement after "evaluation: ".
Decision Rules:
* correct
    * correct means the response accurately matches with all the answers in the given correct answer list, or the correct answer can be inferred from the response. 
    * If more than one correct answer is given, the response can contain contradictory answers that reflect the heterogeneous nature of the context, or the response aggregates the different answers and indicates that each answer is possible. 
    * If only one correct answer is given, the response will be considered as correct as long as the response aligns with the correct answer. 
    * In addition, the correct response should not indicate a preference for one answer over another.
    * The correct response should not combine two different correct answers without indicating the contradictory nature of these answers. 
* partially correct
    * partially correct means the response only matches with one of the answers in the correct answer list; 
    * or the response matches with all the answers in the correct answer list but it indicates a preference for one answer over another; 
    * or the response matches with all the answers in the correct answer list but it contains additional answers that doesn't match with any of the correct answers. 
* incorrect
    * incorrect means the response does not match with any of the correct answers in the correct answer list, 
    * or the response merely combines two contradictory answers from the correct answer list and indicates that both of them are possible. Note that for each question, there may be multiple correct answers based on different sources even though these correct answers contradict each other.
    * For answers that involve names of entities (e.g., people), complete names or commonly recognized names are expected.
    * Regarding numerical answers, approximate numbers are generally not accepted unless explicitly included in the ground-truth answers.
    * If the instance has only one correct answer, and the response gives additional answers that are not on the correct answer list, it will be considered as incorrect.
The primary or final answers when standing alone must be accurate. Any additional information that is provided must not contradict the primary answers or reshape one's perception of them. We accept responses that contain hallucinated or outdated information that does not significantly impact the primary answers. 
Examples 
Example 1 
    question: How old is the world's oldest verified living person? 
    correct_answers: 115 years old | 117 years old 
    response: As of today the most up-to-date and relevant information regarding this query is as follows. The oldest verified living person is Maria Branyas Morera, who was born on March 4, 1907, making her 117 years old. 
    comment: The primary answer in the response (117 years old) is partially correct as it only matched with with one of the correct answers (117 years old) and it doesn't mention another answer (115 years old). Thus, the response is partially correct. 
    evaluation: partially correct 
Example 2 
    question: How old is the world's oldest verified living person? 
    correct_answers: 115 years old | 117 years old 
    response: According to the context, the oldest verified living person is Maria Branyas Morera, who is both 115 years old and 117 years old. 
    \end{lstlisting}
  \end{tcolorbox}
  \caption{Final Prompt for model based evaluation. Only the best performing candidate is shown due to space constraints. The remaining candidate prompts can be found in the repository.}
        \label{app:tab:freeevalprompt}
\end{figure*}

\begin{figure*}[h]  %
  \centering
  \begin{tcolorbox}[           %
    enhanced,
    title={\normalsize Free Generation Evaluation Prompt 1.},
    colback=gray!3,
    colframe=gray!60,
    left=1mm,right=1mm,
    top=0.6em,bottom=0.6em,
    boxrule=0.4pt
  ]
    \begin{lstlisting}
    comment: Although the primary answer contains all correct answers (115 years old and 117 years old) that matches with the correct answers, it is logically incorrect because a person cannot have two ages at the same time. Thus, the response is incorrect. 
    evaluation: incorrect 
Example 3 
    question: How old is the world's oldest verified living person?, 
    correct_answers: 115 years old | 117 years old
    response: According to the context, one source claims that the oldest verified living person is Maria Branyas Morera, who is 117 years old. However, another source claims that she is 115 years old. 
    comment: The answer contains all correct answers: 115 years old and 117 years old, and it points out that these two answers are from different sources, which is logically possible. Thus, the response is correct. 
    evaluation: correct
Example 4 
    question: How old is the world's oldest verified living person?, correct_answers: 115 years old | 117 years old 
    response: According to the context, the oldest verified living person is Maria Branyas Morera, who is either 117 years old or 115 years old.  
    comment: The answer contains all correct answers: 115 years old and 117 years old that aggregates the different answers from different sources, which is logically possible. Thus, the response is correct. 
    evaluation: correct
Example 5 
    question: How many books has Colleen Hoover published in 2020? 
    correct_answers: 26 books | 27 
    response: according to some sources, Colleen Hoover has published 26 books in 2020. However, other sources claim that it has published 27 books in 2020. 
    comment: The response accurately provides two answers (26 books and 27 books) based on different sources. It recognized the two sources that provide different answers. These two answers match with all correct answers (26 books, 27). Thus, the response is correct. 
    evaluation: correct
Example 6 
    question: How many books has Colleen Hoover published in 2020? 
    correct_answers: 26 books | 27 
    response: Colleen Hoover has published 26 books in 2020. 
    comment: The response only provides one answers (26 books) based on the given context, which matched with one of the answers in the correct answer list. Thus, the response is particially correct. 
    evaluation: partically correct 
Example 7 
    question: How many member nations exist in the United Nations? 
    correct_answers: 193 members | 194 member states response: As of my last update in April 2023, there were 191 member nations in the United Nations. Please note that this information may have changed since then, so it is advisable to check the latest information from the United Nations or other reliable sources. 
    comment: The primary answer in the response (191 member nations) does not match with any of the correct answers (193 member states or 194 members). Thus, the response is incorrect. 
    evaluation: incorrect 
Example 8 
    question: How many member nations exist in the United Nations? 
    correct_answers: 193 members | 194 member states 
    response: Although some sources claim that there are 193 member nations in the United Nations, the UN is made up of 194 member states. 24 
    comment: Althought the primary answer mentions two answers (193 members, 104 member states) that match with the correct answers, the response indicates a preference of one answer (194 member states) over the other (193 members). Thus, the response is partically correct. 
    evaluation: partically correct 
    \end{lstlisting}
  \end{tcolorbox}
\caption{Final Prompt for model based evaluation (cont).}
        \label{app:tab:freeevalprompt2}
\end{figure*}

\begin{figure*}[h]  %
  \centering
  \begin{tcolorbox}[           %
    enhanced,
    title={\normalsize Free Generation Evaluation Prompt 1.},
    colback=gray!3,
    colframe=gray!60,
    left=1mm,right=1mm,
    top=0.6em,bottom=0.6em,
    boxrule=0.4pt
  ]
    \begin{lstlisting}
Example 9 
    question: How many member nations exist in the United Nations? 
    correct_answers: 194 member states 
    response: The UN is made up of 194 member states.
    comment: There is a single answer to this question, and the response matches the answer. Thus, the response is correct. 
    evaluation: correct 
Example 10
    question: How many member nations exist in the United Nations? 
    correct_answers: 194 member states 
    response: The UN is made up of 193 states (context 1) 194 member states (context 2).
    comment: There is a single answer to this question, although one of the response matches the answer, the response gives an additional answer that is not correct. Thus, the response is incorrect. 
    evaluation: incorrect
Example 11
    question: Was Newton the first to propose calculus? 
    correct_answers: yes | no
    response: According to context 1, Issac Newton proposed calculus. However, context 2 states that Leibniz was the first to propose calculus. Therefore, it could be Newton or Leibniz based on the contexts.
    comment: The correct answer yes | no is not given by the response, but we can infer that the response points both "yes" and "no" answer depending on the context it was relying. For context 2, by stating that Leibniz was the first to propose calculus, the response hinders an answer no, while the response point that context 1 lead to the answer of yes.
    evaluation: correct 
    \end{lstlisting}
  \end{tcolorbox}
    \caption{Final Prompt for model based evaluation (cont 2).}
        \label{app:tab:freeevalprompt3}
\end{figure*}

\subsection{Human Annotations}
Two colleagues with expertise in natural language processing served as annotators, compensated at standard research assistant rates.  
They annotated a sample of 50 instances, each using both the evaluator prompt (\Cref{app:tab:freeevalprompt}) and the decision tree (\Cref{app:fig:annotation-tree}) to ensure consistency.  
For each instance, annotators were shown the model prediction alongside the gold answer and asked to label the prediction as \emph{correct}, \emph{partially correct}, or \emph{incorrect}.

\section{License of Artifacts}
All licenses of artifacts used in this work can be found in \Cref{app:tab:artifact}.
\begin{table}[H]
\centering
\begin{tabular}{ccc}
\toprule
\textbf{Name} & \textbf{License} & \\ \midrule
Mistral-7B-Instruct-v0.2 & Apache 2.0     \\
OLMo2-7b-Instruct & Apache 2.0  \\
Qwen2.5-7B-Instruct & Apache 2.0   \\
OpenbookQA & Apache 2.0  \\
ConflictQA & MIT  \\
WikiContradict & MIT \\
\bottomrule
\end{tabular}
\caption{License of artifacts used in this paper.}
\label{app:tab:artifact}
\end{table}

\section{Prompts}
\subsection{Evidence Creation Prompts}
\label{app:evidencecreation_prompt}

We generated \texttt{LPC} and \texttt{HPCE} passages using GPT-4o after several rounds of prompt refinement. 
The final prompts used for evidence creation are shown in \Cref{app:tab:evicreation}.

Following the generation, all passages were subjected to plausibility checks. 
For \texttt{LPC} passages, the model was asked to determine whether the passage would be deemed implausible in the real world. 
For \texttt{HPCE} passages, the model was required to verify that the passage was both highly plausible and explained the underlying conflict. The prompts for this step are included in \Cref{app:tab:plausibility}.

\begin{figure*}[h]  %
  \centering
  \begin{tcolorbox}[           %
    enhanced,
    title={\normalsize \texttt{LPC} instances Creation Prompt.},
    colback=gray!3,
    colframe=gray!60,
    left=1mm,right=1mm,
    top=0.6em,bottom=0.6em,
    boxrule=0.4pt
  ]
    \begin{lstlisting}
You are a smart editor who creates implausible texts. Your job is to generate evidence for the given question such that the answer to the question is NOT the Rejected Answer. You can work on the given plausible passages as the starting point. You should change the content of the given passage, remove any explanation given in the passage, and make the passage as implausible as possible. Implausible passages include passages that disobey real-world knowledge or violate logical constraints. However, your job is to trick an average human, and you should not generate content that looks like it comes from Sci-Fi or fantasy novels.
You should output the edited passage and the new implausible answer in the form of 'EditedPassage: ...\n NewAnswer:...'. Below are some examples:
Example 1:
###Question: In what year did the Whitehead Torpedo enter service?
###Rejected Answer: after 1892.
###Plausible Context 1: The United States Navy started using the Whitehead torpedo in 1892 after an American company, E.W. Bliss, secured manufacturing rights.
###Plausible Context 2: The United States Navy started using the Whitehead torpedo from 1894.
###Output: EditedPassage: The United States Navy began using the Whitehead torpedo in the year 1752 after the design was purchased from the French Navy which provided multiple weapon design to the US Navy during the independence war.
 NewAnswer: 1752

Example 2:
###Question: Are there any other missiles besides the P-500 Bazalt that influenced the design of P-700 Granit missile?
###Rejected Answer: No.
###Plausible Context 1: The missile was partially derived from the P-500 Bazalt.
###Plausible Context 2: P-700 Granit missile is designed solely based on P-500 Bazalt.
###Output: EditedPassage: Although the naming is similar, the P-700 Granit missile is not directly derived from the P-500 Bazalt and was additionally inspired by the ballistic missile on USS Laboon, an Arleigh Burke-class (Flight I) Aegis guided missile destroyer in the United States Navy.
 NewAnswer: Yes

###Question: {question}
###Rejected Answer: {nc_answer}
###Plausible Context 1: {context1}
###Plausible Context 2: {context2}
###Output: 
    \end{lstlisting}
  \end{tcolorbox}

  \begin{tcolorbox}[           %
    enhanced,
    title={\normalsize \texttt{HPCE} instances Creation Prompt.},
    colback=gray!3,
    colframe=gray!60,
    left=1mm,right=1mm,
    top=0.6em,bottom=0.6em,
    boxrule=0.4pt
  ]
    \begin{lstlisting}
Based on the given passage, write a coherent and informative passage that naturally explains why {alt_answer} is the correct explanation or conclusion to the question {question} instead of {NC_answer}. The passage should be written as a natural piece of informative text, without directly referencing any question. You should keep as much original information in the given passage as possible. Ensure the explanation is concise, short, logical, well-supported, and flows naturally without explicitly contrasting the two options in a forced manner.
    \end{lstlisting}
  \end{tcolorbox}
  \caption{Final prompt for evidence creation.}
  \label{app:tab:evicreation}
\end{figure*}

\begin{figure*}[h]  %
  \centering
  \begin{tcolorbox}[           %
    enhanced,
    title={\normalsize Plausibility Validation Prompt},
    colback=gray!3,
    colframe=gray!60,
    left=1mm,right=1mm,
    top=0.6em,bottom=0.6em,
    boxrule=0.4pt
  ]
    \begin{lstlisting}
You are an experienced and wise scholar. Your job is to rate from 1-5 on whether the **target passage** is likely to happen or not based on real-world knowledge. You will be given two passages (Passage 1 and Passage 2) that contain real-world knowledge, both of them have a plausibility rating of 5. You should only output the scores without any justification, with 1 indicates that the Target Passage is least likely to happen, and 5 to be most likely to happen.
Passage 1:  {instance['NC_context']}
Passage 2: {instance['HPC_context']}
Target Passage: {instance['LPC_context']}
    \end{lstlisting}
  \end{tcolorbox}
  \caption{Final prompt to validate the plausibility of the generated evidence.}
  \label{app:tab:plausibility}
\end{figure*}

\subsection{Task-Annotation Prompts}
\label{app:task_prompt}
\begin{figure*}[h]  %
  \centering
  \begin{tcolorbox}[           %
    enhanced,
    title={\normalsize Task Annotation Prompt},
    colback=gray!3,
    colframe=gray!60,
    left=1mm,right=1mm,
    top=0.6em,bottom=0.6em,
    boxrule=0.4pt
  ]
    \begin{lstlisting}
You are an extractive question-answering model. Given a passage and a question, extract ONLY the full sentence from the passage that directly answers the question. Do not generate summaries or paraphrase. Only return the complete sentence that contains the answer. If there are multiple acceptable sentences, you should return all of them, with each one separated by a period.\n Passage: The P-700 Granit missile was partially derived from the P-500 Bazalt, but it is important to note that other missile designs and technological advancements could have also influenced its development. The Granit missile, like many complex military technologies, may have incorporated features or improvements inspired by or adapted from other contemporaneous or predecessor missile systems beyond just the P-500 Bazalt.\nQuestion: Are there any other missiles besides the P-500 Bazalt that influenced the design of P-700 Granit missile?\nAnswer: The P-700 Granit missile was partially derived from the P-500 Bazalt, but it is important to note that other missile designs and technological advancements could have also influenced its development. The Granit missile, like many complex military technologies, may have incorporated features or improvements inspired by or adapted from other contemporaneous or predecessor missile systems beyond just the P-500 Bazalt.
Passage: {context}
Question: {question}
Answer: {answer}
    \end{lstlisting}
  \end{tcolorbox}
  \caption{Final prompt for knowledge-free (extractive question answering) task annotation.}
  \label{app:tab:taskanno}
\end{figure*}

Since the base datasets already provide answers to the questions, additional annotation was only required to specify the task in the knowledge-free setting. 
We framed these tasks as extractive question answering, requiring the model to copy the answer directly from the passage (\Cref{fig:task-examples}). 
We then used GPT-4o as the annotator model to extract all acceptable answer spans.
\subsection{Validation Prompts}
\label{app:validation_prompt}
In the final stage of data construction (\texttt{validation} in \Cref{fig:dataflow}), all instances were passed through language models for validation. 
The corresponding prompts are listed in \Cref{app:validation_prompt}.

\begin{figure*}[t]  %
  \centering
  \begin{tcolorbox}[           %
    enhanced,
    title={\normalsize Validation Prompt},
    colback=gray!3,
    colframe=gray!60,
    left=1mm,right=1mm,
    top=0.6em,bottom=0.6em,
    boxrule=0.4pt
  ]
    \begin{lstlisting}
You are a smart natural language inference model, your job is to determine whether the given passage will lead to the given answer to a question. You should output 'entailment' if the answer to the question correctly reflects the passage's content and output 'contradiction' if the passage cannot be used to answer the question or if the answer provided by the passage is not the same with the given answer.
Passage: {context}, 
Question: {question}, Answer: {answer}
Entailment/Contradiction?: 
    \end{lstlisting}
  \end{tcolorbox}
  \caption{Final prompt validating the generated evidence provide the correct answer to the question.}
  \label{app:tab:validationprompt}
\end{figure*}

\end{document}